\documentclass[runningheads]{llncs}
\usepackage{graphicx}
\usepackage{comment}
\usepackage{amsmath,amssymb} % define this before the line numbering.
\usepackage{color}
\usepackage{graphicx}
\usepackage{float}
\usepackage{amsmath}
\usepackage{amsfonts}
\usepackage{amssymb}
\usepackage{multirow}
\usepackage{tabu}
\usepackage[title]{appendix}
\usepackage[english]{babel}
\usepackage{bm}
\usepackage{colortbl}
\usepackage{enumitem}
\usepackage{array}
\usepackage[ruled,vlined]{algorithm2e}
\usepackage{threeparttable}
\usepackage{dsfont}
\usepackage{pifont}
\usepackage{subcaption}
\usepackage{nccmath}
\usepackage{booktabs}
\usepackage{listings,lstautogobble}
\usepackage{marvosym}
\usepackage{pbox}
\usepackage{tabularx}
\usepackage[super]{nth}

\newcolumntype{Y}{>{\centering\arraybackslash}X}
\newcolumntype{C}[1]{>{\centering\arraybackslash}p{#1}}

\newlength\savewidth\newcommand\shline{\noalign{\global\savewidth\arrayrulewidth
  \global\arrayrulewidth 1pt}\hline\noalign{\global\arrayrulewidth\savewidth}}
\newcolumntype{x}[1]{>{\centering\arraybackslash}p{#1pt}}
\newcommand{\tablestyle}[2]{\setlength{\tabcolsep}{#1}\renewcommand{\arraystretch}{#2}\centering\small}
\newcommand{\bd}[1]{\textbf{#1}}
\newcolumntype{Y}{>{\centering\arraybackslash}X}

\begin{document}
\pagestyle{headings}
\mainmatter

\title{SegFix: Model-Agnostic Boundary Refinement for Segmentation}

\titlerunning{SegFix: Model-Agnostic Boundary Refinement for Segmentation}
\authorrunning{Yuhui Yuan, Jingyi Xie, Xilin Chen, Jingdong Wang}

\author{Yuhui Yuan\inst{1,2,4\star} \and
Jingyi Xie$^{3}$\thanks{Equal contribution.} \and
Xilin Chen\inst{1,2} \and
Jingdong Wang\inst{4}
}
\institute{
\footnotesize
Key Lab of Intelligent Information Processing of Chinese Academy of Sciences (CAS), Institute of Computing Technology, CAS \and
University of Chinese Academy of Sciences \and
University of Science and Technology of China \and
Microsoft Research Asia \\
\email{\{yuhui.yuan, jingdw\}@microsoft.com, hsfzxjy@mail.ustc.edu.cn, xlchen@ict.ac.cn}}

%******************
\maketitle

\vspace{-4mm}
\begin{abstract}
We present a model-agnostic post-processing scheme to improve the boundary quality for the segmentation 
result that is generated by any existing segmentation model.
Motivated by the empirical observation that the label predictions of interior pixels are more reliable, 
we propose to replace the originally unreliable predictions
of boundary pixels by the predictions of interior pixels. Our approach processes only the input image 
through two steps: (i) localize the boundary pixels
and (ii) identify the corresponding interior pixel for each boundary pixel. 
We build the correspondence by learning a direction away from the boundary pixel
to an interior pixel. Our method requires no prior information of the segmentation models 
and achieves nearly real-time speed. We empirically verify that our SegFix consistently 
reduces the boundary errors for segmentation results generated from various state-of-the-art 
models on Cityscapes, ADE20K and GTA5. Code is available at:
{\url{\color{blue}{https://github.com/openseg-group/openseg.pytorch}}}.
\end{abstract}

%%%%%%%%%%%%%%%%%%%%%%%%%%%%%%%%%%%%%%%%%%%%%%%%%%%%%%%%%%%
\section{Introduction}
\label{sec:introduction}
\vspace{-3mm}
%%%%%%%%%%%%%%%%%%%%%%%%%%%%%%%%%%%%%%%%%%%%%%%%%%%%%%%%%%%

The task of semantic segmentation is formatted as predicting 
the semantic category for each pixel in an image.
Based on the pioneering fully convolutional network~\cite{long2015fully},
previous studies have achieved great success as reflected by increasing the performance on various challenging semantic segmentation benchmarks~\cite{caesar2018coco,cordts2016cityscapes,zhou2017scene}.

{
\begin{figure}
\centering
\normalsize
\begin{minipage}[c]{0.6\textwidth}
{
	\includegraphics[width=.2\textwidth]{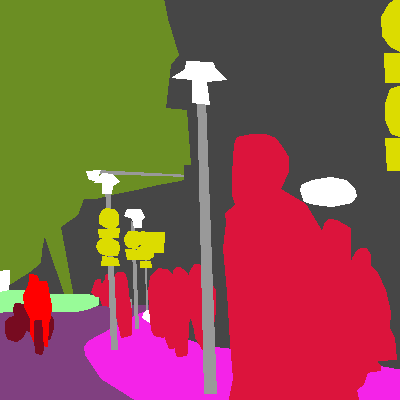}
	\frame{\includegraphics[width=.2\textwidth]{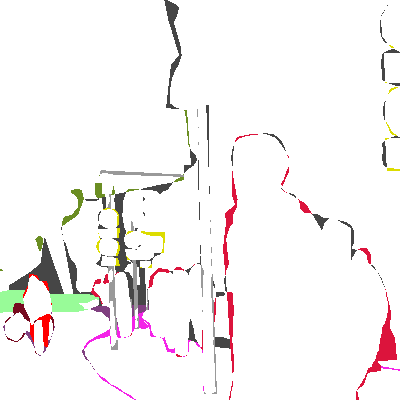}}
	\frame{\includegraphics[width=.2\textwidth]{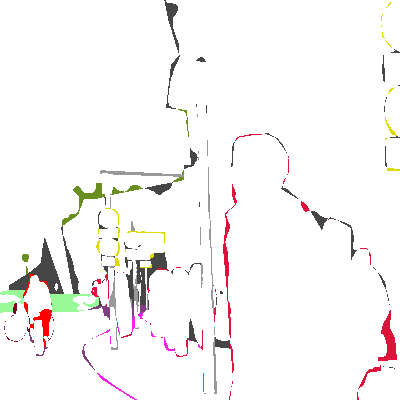}}
	\frame{\includegraphics[width=.2\textwidth]{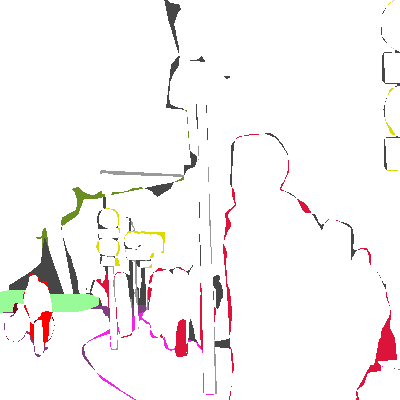}} \\
% 	\vspace{.05cm}
	{\includegraphics[width=.2\textwidth]{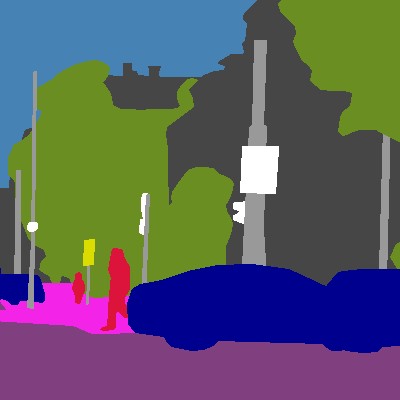}}
	\frame{\includegraphics[width=.2\textwidth]{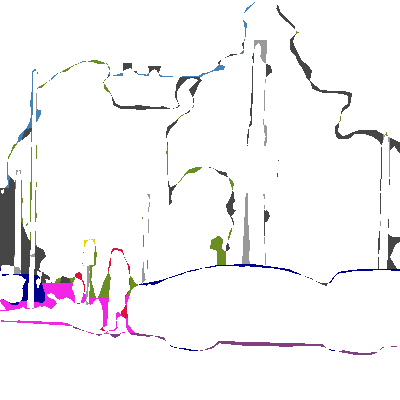}}
	\frame{\includegraphics[width=.2\textwidth]{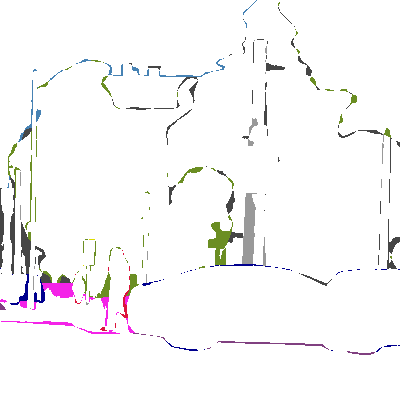}}
	\frame{\includegraphics[width=.2\textwidth]{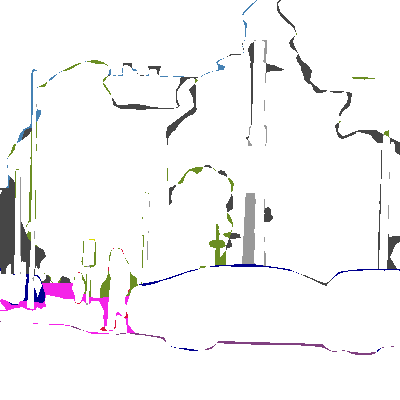}} \\
% 	\vspace{.05cm}
	{\includegraphics[width=.2\textwidth]{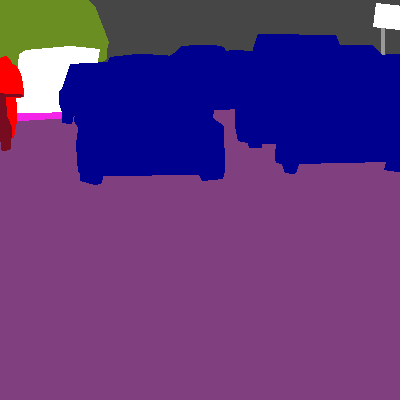}}
	\frame{\includegraphics[width=.2\textwidth]{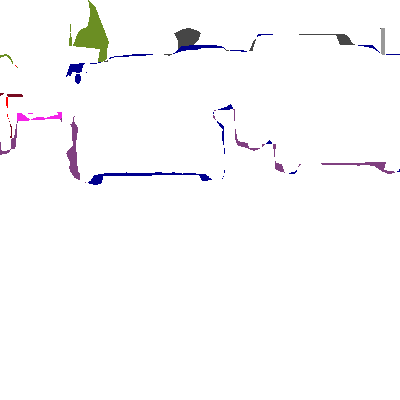}}
	\frame{\includegraphics[width=.2\textwidth]{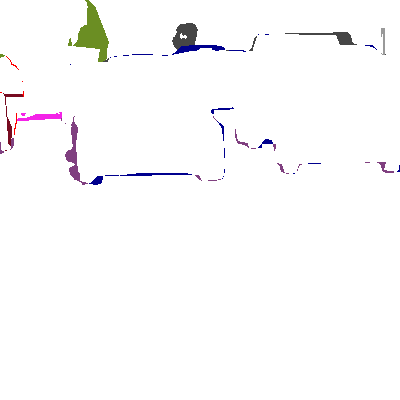}}
	\frame{\includegraphics[width=.2\textwidth]{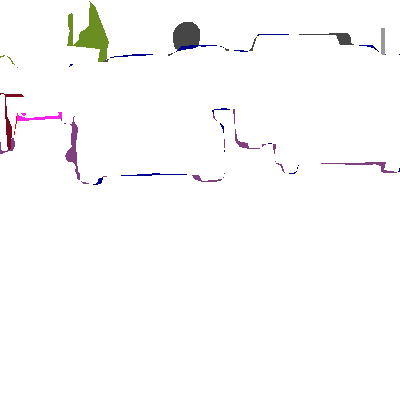}} \\
}
\end{minipage}
\hspace{-7mm}
\begin{minipage}[c]{0.43\textwidth}
\caption{\small{
{\textbf{Qualitative analysis of the segmentation error maps.}}
The \nth{1} column presents the ground-truth segmentation maps, 
and the \nth{2}/\nth{3}/\nth{4} column presents the 
the error maps of DeepLabv3/HRNet/Gated-SCNN separately.
These examples are cropped from Cityscapes \texttt{val} set.
We can see that there exist many errors along the thin boundary for all three methods.}}
\label{fig:boundary_pixel_example}
\end{minipage}
\vspace{-6mm}
\end{figure}
}

Most of the existing works mainly addressed semantic segmentation
through (i) increasing the resolution of feature maps~\cite{chen2017rethinking,chen2018encoder,sun2019high}, (ii) constructing
more reliable context information~\cite{zhao2017pyramid,yuan2018ocnet,fu2019dual,huang2019interlaced,zhang2019co}
and (iii) exploiting boundary information~\cite{bertasius2016semantic,chen2016semantic,liu2018devil,takikawa2019gated}.
In this work,
we follow the \nth{3} line of work and focus on improving segmentation
result on the pixels located within the thinning boundary\footnote{
In this paper,
we treat the pixels with neighboring pixels belonging to different 
categories as the boundary pixels.
We use the distance transform to generate the ground-truth boundary map
with any given width in our implementation.
}
via a simple but effective
model-agnostic boundary refinement mechanism.

\begin{figure}
\centering
\hspace{-10mm}
\begin{subfigure}[b]{0.32\textwidth}
{\includegraphics[width=\textwidth]{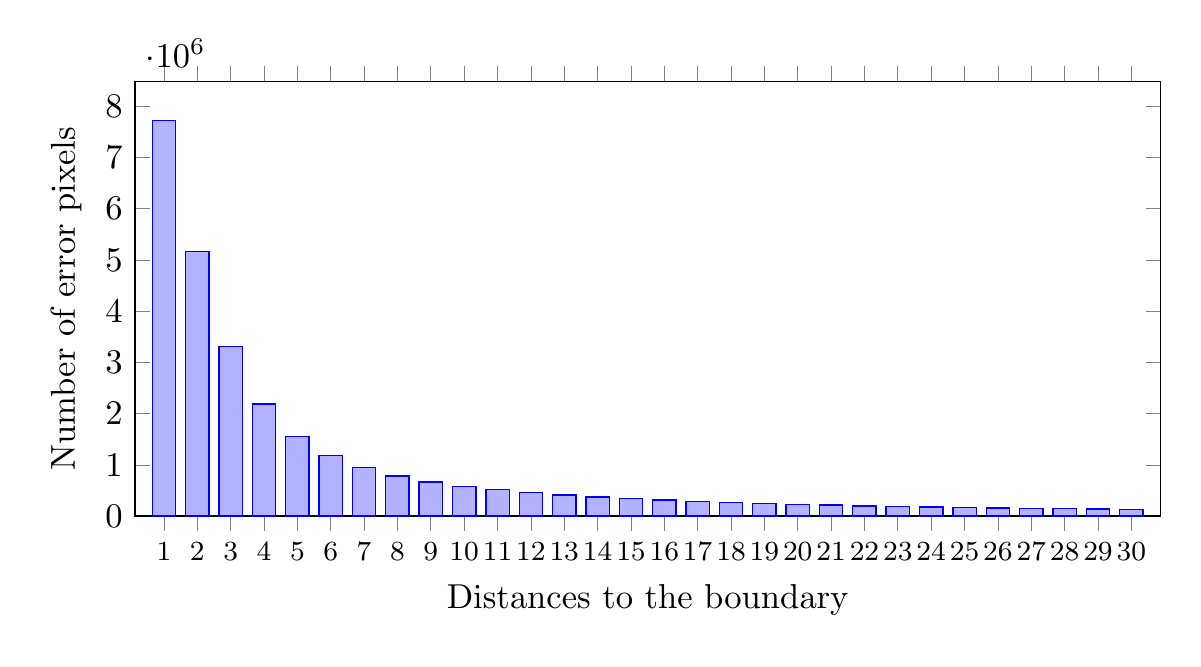}}
\caption{DeepLabv3~\cite{chen2017rethinking}}
\end{subfigure}
\begin{subfigure}[b]{0.32\textwidth}
{\includegraphics[width=\textwidth]{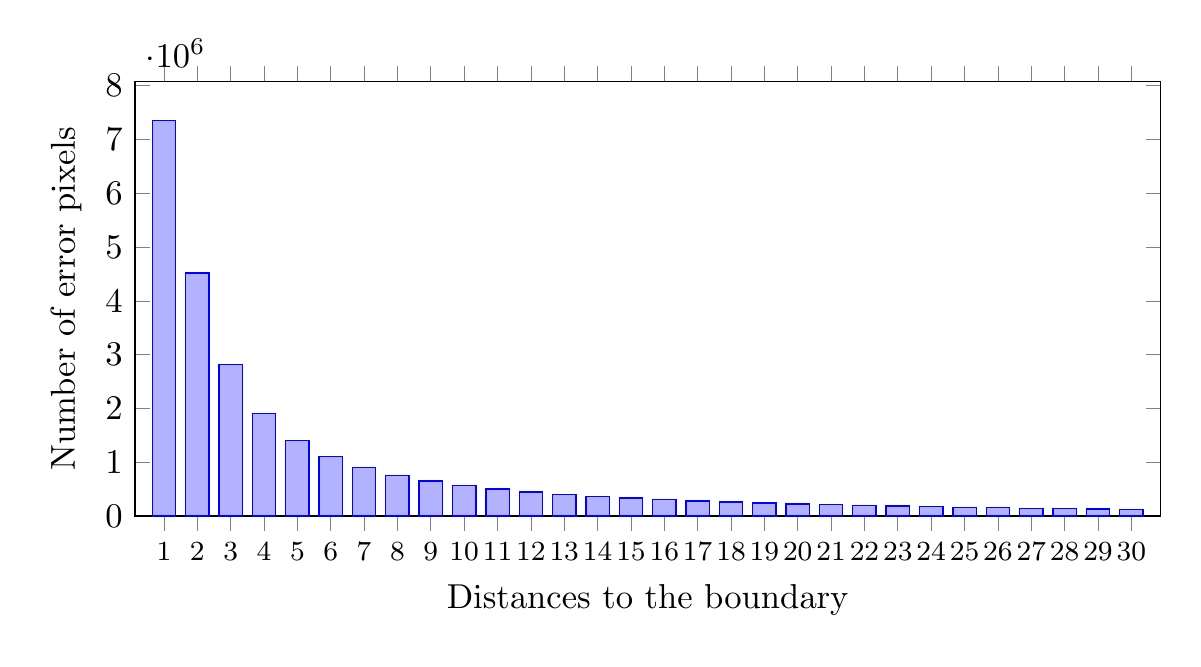}}
\caption{\small{HRNet~\cite{sun2019high}}}
\end{subfigure}
\begin{subfigure}[b]{0.32\textwidth}
{\includegraphics[width=\textwidth]{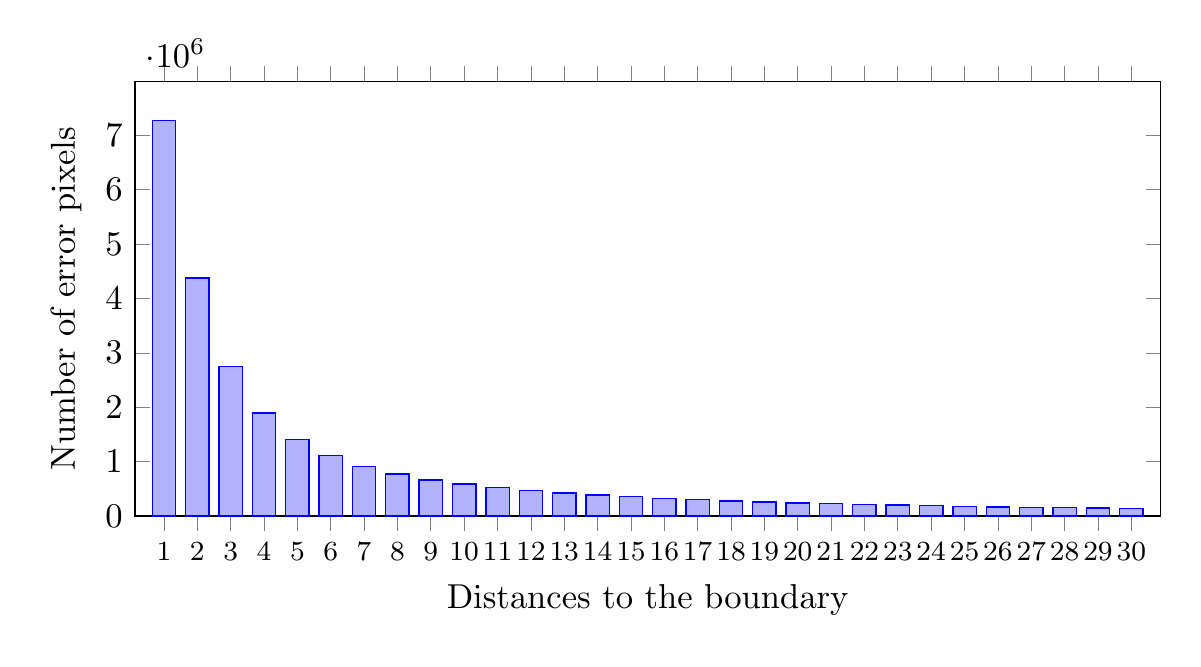}}
\caption{Gated-SCNN~\cite{takikawa2019gated}}
\end{subfigure}
\hspace{-6mm}
\caption{
\small{
\textbf{Histogram statistics of errors:}
the number of error pixels \emph{vs.} their (Euclidean) distances to the boundaries on Cityscapes \texttt{val} based on DeepLabv3/HRNet/Gated-SCNN.
We can see that pixels with larger distance tend to be well-classified with higher probability and 
there exist many errors distributing within $\sim5$ pixels width along the boundary.
}
}
\label{fig:boundary_distance_map}
\vspace{-5mm}
\end{figure}

Our work is mainly motivated by the observation that
\emph{most of the existing state-of-the-art segmentation models
fail to deal well with the error predictions along the boundary.}
We illustrate some examples of the segmentation error maps with DeepLabv3~\cite{chen2017rethinking},
Gated-SCNN~\cite{takikawa2019gated} and HRNet~\cite{sun2019high}
in Figure~\ref{fig:boundary_pixel_example}.
More specifically, we illustrate the statistics on the 
numbers of the error pixels \emph{vs.} the distances
to the object boundaries in Figure~\ref{fig:boundary_distance_map}.
We can observe that, for all three methods, the number of error pixels significantly decrease 
with larger distances to the boundary.
In other words, 
predictions of the interior pixels are more reliable.

We propose a novel model-agnostic post-processing
mechanism to reduce boundary errors by 
replacing labels
of boundary pixels with the labels of 
corresponding interior pixels
for a segmentation result.
We estimate the pixel correspondences by processing the input image
(without exploring the segmentation result)
with two steps.
The first step aims to 
localize the pixels along the object boundaries.
We follow the contour detection methods~\cite{Bertasius_2015_ICCV,arbelaez2010contour,DollarARXIV14edges}
and simply use a convolutional network
to predict a binary mask indicating the boundary pixels.
In the second step,
we learn a direction away from the boundary pixel to an interior pixel
and identify the corresponding interior pixel
by moving from the boundary pixel
along the direction 
by a certain distance.
Especially,
our SegFix can reach nearly real-time speed with high resolution inputs.

Our SegFix is a general scheme that consistently 
improves the performance
of various segmentation models
across multiple benchmarks without any prior information.
We evaluate the effectiveness of SegFix
on multiple semantic segmentation benchmarks
including Cityscapes, ADE20K and GTA5.
We also extend SegFix to
instance segmentation task on Cityscapes.
According to the Cityscapes leaderboard,
``HRNet + OCR + SegFix" and ``PolyTransform + SegFix" achieve
$84.5\%$ and $41.2\%$, which rank the \nth{1} and \nth{2} place
on the semantic and instance segmentation track separately
by the ECCV 2020 submission deadline.

%%%%%%%%%%%%%%%%%%%%%%%%%%%%%%%%%%%%%%%%%%%%%%%%%%%%%%%%%%%
\vspace{-1mm}
\section{Related Work}
\vspace{-1mm}
%%%%%%%%%%%%%%%%%%%%%%%%%%%%%%%%%%%%%%%%%%%%%%%%%%%%%%%%%%%

\noindent \textbf{Distance/Direction Map for Segmentation:}
Some recent work~\cite{bai2017deep,hayder2017boundary,wang2019object}
performed distance transform to compute distance maps for instance segmentation task.
For example, 
\cite{bai2017deep,hayder2017boundary} proposed to train the model to 
predict the truncated distance maps within each cropped instance.
The other work~\cite{dangi2019distance,bischke2019multi,chen2018masklab,papandreou2018personlab}
proposed to regularize the semantic or instance segmentation predictions
with distance map or direction map in a multi-task mechanism.
Compared with the above work,
the key difference is that 
our approach does not perform any segmentation predictions
and instead predicts the direction map from only the image, 
and then we refine the segmentation results of the existing approaches.

\noindent \textbf{Level Set for Segmentation:} Many previous efforts ~\cite{osher1988fronts,caselles1997geodesic,kass1988snakes} have used 
the level set approach to
address the semantic segmentation problem
before the era of deep learning.
The most popular formulation of level set is
the signed distance function,
with all the zero values corresponding to predicted boundary positions.
Recent work~\cite{acuna2019devil,chen2019learning,wang2019object,kim2019cnn} extended the conventional level-set scheme to deep network for regularizing the boundaries of predicted segmentation map.
Instead of representing the boundary with a level set function directly,
we implicitly encode the relative 
distance information of the boundary pixels
with a boundary map and a direction map.

\noindent \textbf{DenseCRF for Segmentation:}
Previous work~\cite{chen2017deeplab,yu2015multi,Zheng_2015_ICCV,Lin_2017_CVPR} improved their segmentation results
with the DenseCRF~\cite{krahenbuhl2011efficient}.
Our approach is also a kind of general post processing scheme
while being simpler and more efficient for usage.
We empirically show that our approach
not only outperforms but also is complementary with 
the DenseCRF.

\noindent \textbf{Refinement for Segmentation:}
Extensive studies~\cite{fieraru2018learning,gidaris2017detect,islam2017label,kuo2019shapemask,li2016iterative} have proposed various mechanisms to
refine the segmentation maps from coarse to fine.
Different from most of the existing refinement approaches
that depend on the segmentation models,
to the best of our knowledge,
our approach is the first model-agnostic segmentation refinement
mechanism that can be applied to refine the segmentation results of 
any approach without any prior information.

{
\begin{figure*}[t]
\centering
\includegraphics[width=0.96\textwidth]{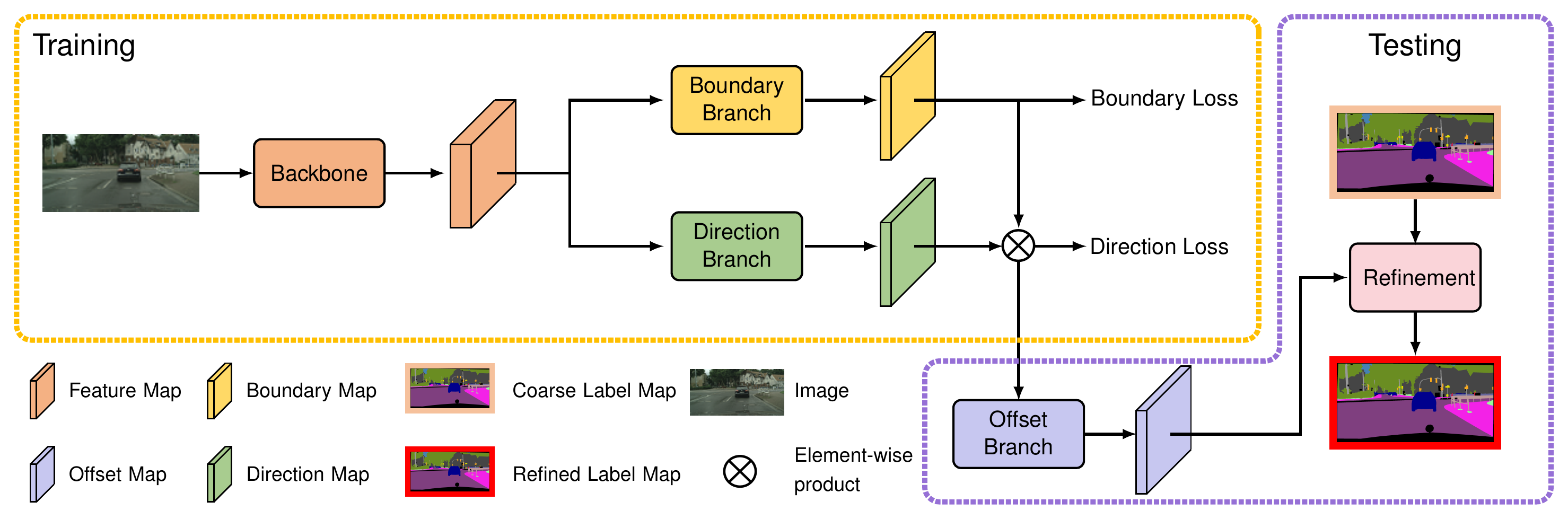}
\vspace{-4mm}
\caption{\small{
\textbf{
Illustrating the SegFix framework:
}
In the training stage,
we first send the input image into a backbone to predict a feature map.
Then we apply a boundary branch to predict a binary boundary map 
and a direction branch to predict a direction map and
mask it with the binary boundary map. 
We apply boundary loss and direction loss 
on the predicted boundary map and direction map separately.
In the testing stage,
we first convert the direction map to offset map
and then refine the segmentation results of any existing methods 
according to the offset map.
}}
\vspace{-6mm}
\label{fig:pipeline}
\end{figure*}
}

\noindent \textbf{Boundary for Segmentation:}
Some previous efforts~\cite{acuna2019devil,yu2017casenet,liu2017richer,yu2018simultaneous} focused on localizing semantic boundaries.
Other studies~\cite{bertasius2016semantic,takikawa2019gated,ding2019boundary,liu2018devil,liu2017learning,ke2018adaptive,ding2019semantic,ding2018context} also exploited the boundary information to improve the segmentation. 
For example,
BNF~\cite{bertasius2016semantic} introduced a global energy model to
consider the pairwise pixel affinities based on the boundary predictions. 
Gated-SCNN~\cite{takikawa2019gated} exploited the
duality between the segmentation predictions and the
boundary predictions with a two-branch mechanism and a regularizer.

These methods~\cite{bertasius2016semantic,ding2019boundary,takikawa2019gated,ke2018adaptive} 
are highly dependent on the segmentation models
and require careful re-training or fine-tuning.
Different from them,
SegFix does not perform either segmentation prediction or 
feature propagation and we instead refine the 
segmentation maps with an offset map
directly.
In other words,
we only need to train a single unified SegFix model once w/o any
further fine-tuning the different segmentation models (across multiple different datasets).
We also empirically verify that our approach is complementary with the above methods,
e.g., Gated-SCNN~\cite{takikawa2019gated} and Boundary-Aware Feature Propagation~\cite{ding2019boundary}.

\noindent \textbf{Guided Up-sampling Network:}
The recent work~\cite{mazzini2018guided,Mazzini_2019_CVPR_Workshops}
performed a segmentation guided offset scheme 
to address boundary errors caused by the bi-linear up-sampling.
The main difference is that they do not apply any explicit supervision on their offset maps and require re-training for different models, while 
we apply explicit semantic-aware supervision on the offset maps and 
our offset maps can be applied to
various approaches directly without any
re-training.
We also empirically verify the advantages of our approach.

\noindent \textbf{Semantically Thinned Edge Alignment Learning (STEAL):}
The previous study STEAL~\cite{acuna2019devil} is the most similar work
as it also predicts both boundary maps and direction maps (simultaneously) to refine the 
boundary segmentation results. 
To justify the main differences between STEAL and our SegFix, 
we summarize several key points as following: 
(i) STEAL predicts $K$ independent boundary maps (associated with $K$ categories) 
while SegFix only predicts a single boundary map w/o differentiating the different categories.
(ii) STEAL first predicts the boundary map and then applies a fixed convolution on the boundary map to estimate the direction map while SegFix uses two parallel branches to predict them independently.
(iii) STEAL uses mean-squared-loss on the direction branch while SegFix uses cross-entropy loss (on the discrete directions).
Besides,
we empirically compare STEAL and our SegFix in the ablation study.

%%%%%%%%%%%%%%%%%%%%%%%%%%%%%%%%%%%%%%%%%%%
\vspace{-3mm}
\section{Approach}
\label{methods}
\vspace{-2mm}
%%%%%%%%%%%%%%%%%%%%%%%%%%%%%%%%%%%%%%%%%%%

\subsection{Framework}
The overall pipeline of 
SegFix is illustrated in Figure~\ref{fig:pipeline}.
We first train a model to pick out boundary pixels 
(with the boundary maps) and
estimate their corresponding interior pixels
(with offsets derived from the direction maps)
from only the image.
We do not perform segmentation 
directly during training.
We apply this model to generate offset maps from the images
and use the offsets to get the corresponding pixels which should
mostly be the more confident interior pixels,
and thereby 
refine segmentation results 
from any segmentation model.
We mainly describe SegFix scheme for semantic segmentation
and we illustrate the details for instance segmentation in the Appendix.

% \vspace{.1cm}
\noindent\textbf{Training stage.}
Given an input image $\mathbf{I}$ of shape $H\times W\times 3$, 
we first use a backbone network to 
extract a feature map $\mathbf{X}$,
and then send $\mathbf{X}$ in parallel to
(1) the \emph{boundary branch} to predict
a binary map $\mathbf{B}$,
with $1$ for the boundary pixels
and $0$ for the interior pixels,
and 
(2) the \emph{direction branch}
to predict a direction map $\mathbf{D}$
with each element storing the direction 
pointing from the boundary pixel to the interior pixel.
The direction map $\mathbf{D}$ is then
masked by the binary map $\mathbf{B}$
to yield the input for the offset branch.

For model training, we use a binary cross-entropy loss
as the boundary loss
on $\mathbf{B}$ and a categorical cross-entropy loss
as the direction loss on
$\mathbf{D}$ separately.

% \vspace{.1cm}
\noindent\textbf{Testing stage.}
Based on the predicted boundary map $\mathbf{B}$ and direction map $\mathbf{D}$, 
we apply the \emph{offset branch} to generate a offset map $\Delta{\mathbf{Q}}$.
A coarse label map ${\mathbf{L}}$ output by any semantic segmentation model
will be refined as:

\begin{ceqn}
\begin{align}
\widetilde{\mathbf{L}}_{\mathbf{p}_i} &= {\mathbf{L}}_{\mathbf{p}_i+\Delta{\mathbf{q}_i}},
\end{align}
\end{ceqn}

\noindent where $\widetilde{\mathbf{L}}$ is refined label map,
$\mathbf{p}_i$ represents the coordinate of the boundary pixel $i$, $\Delta{\mathbf{q}_i}$ is the generated offset vector pointing to an interior pixel,
which is indeed an element of $\Delta{\mathbf{Q}}$.
$\mathbf{p}_i+\Delta{\mathbf{q}_i}$ is the position of 
the identified interior pixel.

Considering that there might be some ``fake" interior pixels
\footnote{
We use ''fake" interior pixels to represent
pixels (after offsets)
that still lie on the boundary
when the boundary is thick.
Notably, we identify an pixel as interior pixel/boundary pixel if its value in the predicted boundary map $\mathbf{B}$ is $0$/$1$.
% These ''fake" interior pixels can still be helpful for refinement as
% long as they are closer to the inside of the corresponding object.
}
when the boundary is thick,
we propose two different schemes as following:
(i) re-scaling all the offsets by a factor, e.g., 2.
(ii) iteratively applying the offsets (of the ``fake" interior pixels) until finding an interior pixel.
We choose (i) by default for simplicity as their performance is close.

During testing stage,
we only need to generate the offset maps on test set
\emph{for once},
and could 
apply the same offset maps to refine the segmentation results from
any existing segmentation model without requiring any prior information.
In general,
our approach is agnostic to any existing segmentation models.

{
\begin{figure*}[t]
\centering
\includegraphics[width=0.95\textwidth]{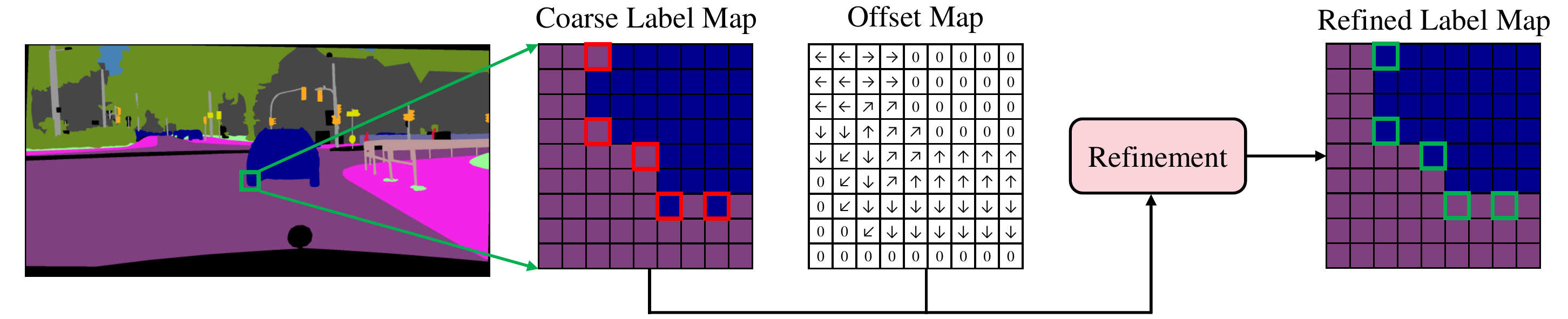}
\caption{\small{
\textbf{Illustrating the refinement mechanism of our approach:}
we refine the coarse label map based on the offset map by 
replacing the labels of boundary pixels with the labels of 
(more) reliable interior pixels.
We represent different offset vectors with different arrows.
We mark the error positions in the coarse label map with $\textcolor{red}{\Box}$
and the corresponding corrected positions in the refined label map with $\textcolor{green}{\Box}$.
For example, the top-most error pixel (class road) in the coarse label map is associated with a direction $\rightarrow$.
We use the label (class car) of the updated position based on offset ($1,0$)
as the refined label.
Only one-step shifting based on the offset
map already refines several boundary errors.
}}
\label{fig:approach:shift-improve}
\vspace{-8mm}
\end{figure*}
}

\vspace{-3mm}
\subsection{Network Architecture}
% \vspace{.1cm}
\noindent\textbf{Backbone.}
We adopt the recently proposed
high resolution network (HRNet)~\cite{sun2019high} 
as backbone,
due to its strengths at maintaining high resolution feature maps
and 
our need to apply full-resolution boundary maps and direction maps 
to refine full-resolution coarse label maps.
To further increase the resolution of the output feature map,
we modify the original HRNet through
adding an extra branch that maintains higher output resolution, i.e., $2\times$,
% on the final output feature map of HRNet
% to increase the resolution by $2\times$,
% which is similar to~\cite{cheng2019bottom},
called HRNet-${2\times}$.
We directly perform the boundary branch and the direction branch
on the output feature map with the highest resolution.
The resolution is $\frac{H}{s}\times \frac{W}{s}\times D$,
where $s=4$ for HRNet and $s=2$ for HRNet-${2\times}$.
We empirically verify that our approach consistently 
improves the coarse segmentation results
for all variations of our backbone choices in Section~\ref{sec:ablation},
e.g., HRNet-W$18$ and HRNet-W$32$.

% \vspace{.1cm}
\noindent\textbf{Boundary branch/loss.}
We implement the boundary branch as
$1\times 1~\operatorname{Conv} \rightarrow \operatorname{BN} \rightarrow \operatorname{ReLU}$ with $256$ output channels.
We then apply a linear classifier 
($1\times 1~\operatorname{Conv}$) and up-sample 
the prediction to generate the final
boundary map $\mathbf{B}$ of size $H\times W\times 1$.
Each element of $\mathbf{B}$ records
the probability of the pixel belonging to the boundary.
We use binary cross-entropy loss as the boundary loss.

% \vspace{.1cm}
\noindent\textbf{Direction branch/loss.}
Different from the previous approaches~\cite{acuna2019devil,bai2017deep}
that perform regression on continuous directions in $[0^{\circ}, 360^{\circ})$ as 
the ground-truth,
our approach directly predicts discrete directions
by evenly dividing the entire direction range
to $m$ partitions (or categories) as our ground-truth ($m=8$ by default).
In fact, we empirically find that our discrete categorization scheme
outperforms the regression scheme, e.g., mean squared loss in the angular domain~\cite{bai2017deep}, measured by the final segmentation performance improvements.
We illustrate more details for the discrete direction map in Section~\ref{sec:gtdir}.

\begin{figure*}
\centering
\begin{subfigure}[b]{0.45\textwidth}
\vspace{-2mm}
{\includegraphics[width=\textwidth]{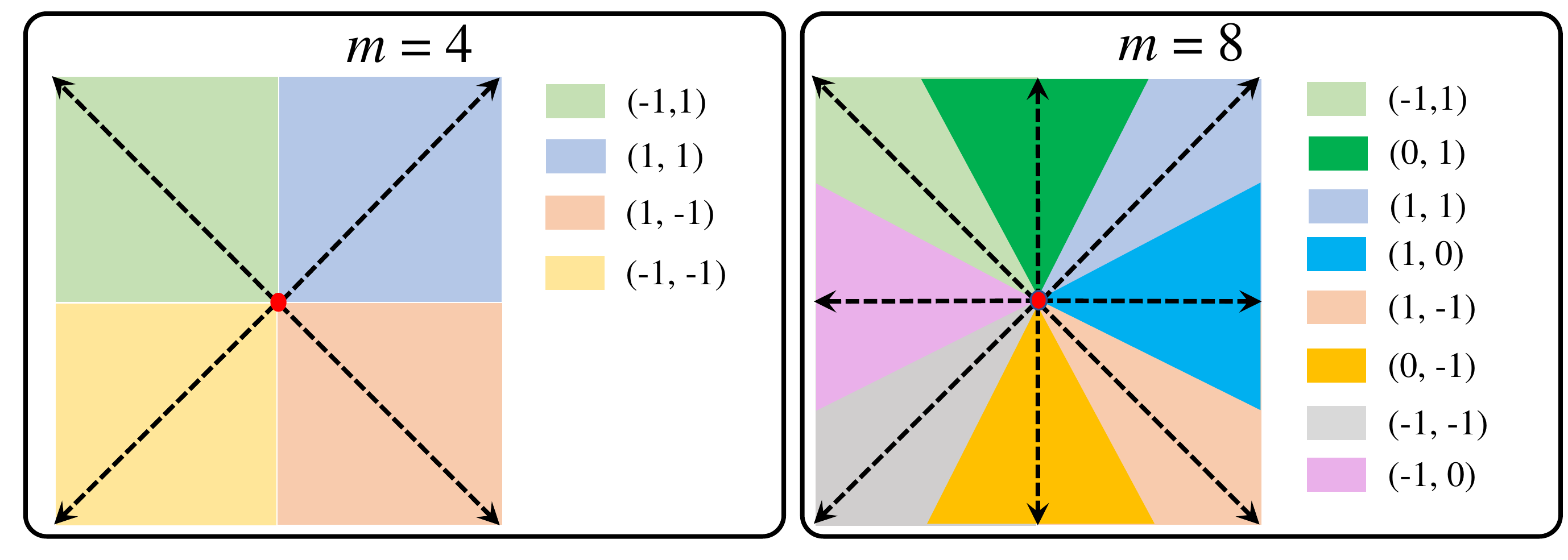}}
\caption{Illustrating the directions $\to$ offsets.}
\end{subfigure}
\hspace{3mm}
\begin{subfigure}[b]{0.47\textwidth}
{\includegraphics[width=\textwidth]{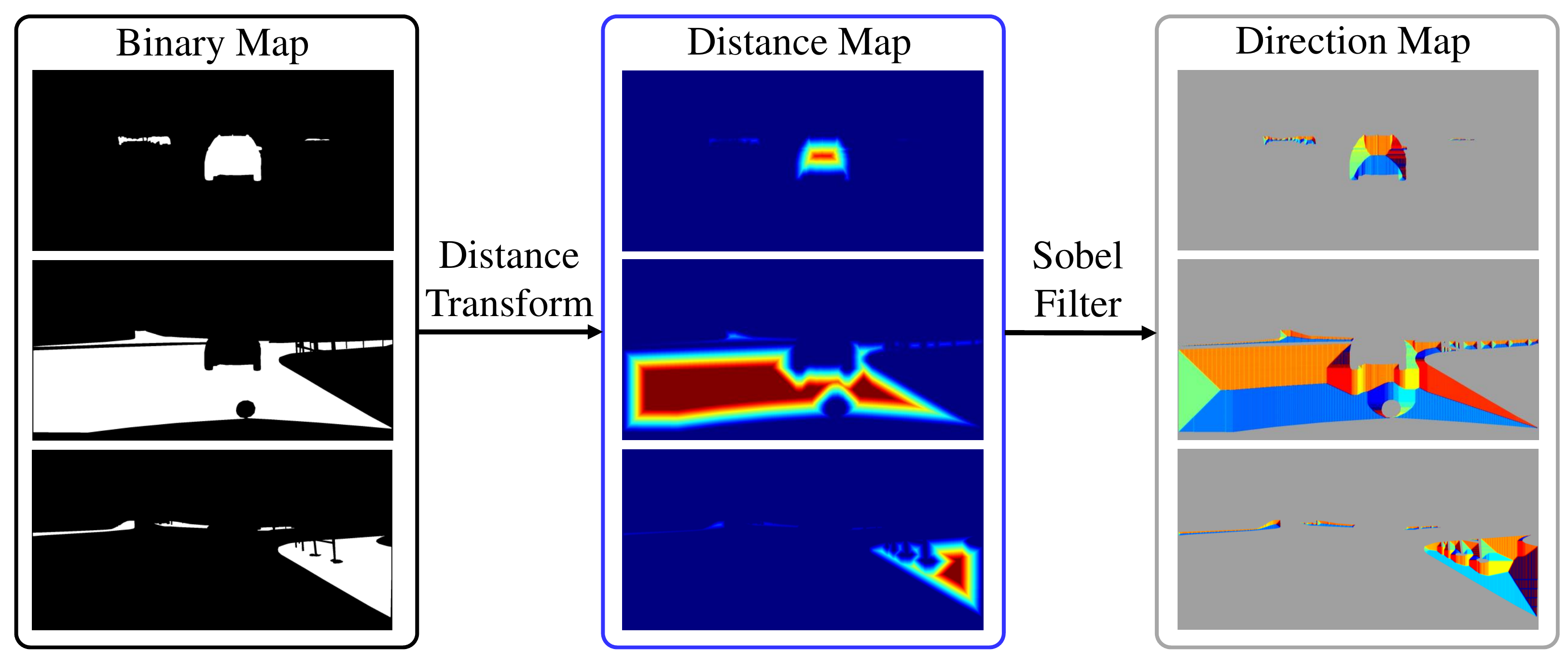}}
\caption{Ground-truth generation procedure.}
\end{subfigure} \\
\vspace{-3mm}
\caption{
\small{
(a) We divide the entire direction value range
$[0^{\circ}, 360^{\circ})$
% based on the positive $x$-axis) 
to $m$ partitions or categories (marked with different colors), 
For example, when $m=4$, 
we have $[0^{\circ}, 90^{\circ})$, $[90^{\circ}, 180^{\circ})$, $[180^{\circ}, 270^{\circ})$ and $[270^{\circ}, 360^{\circ})$
correspond to $4$ different categories separately. 
The above $4$ direction categories
correspond to offsets $(1,1)$, $(-1,1)$, $(-1,-1)$
and $(1,-1)$ respectively. 
The situation for $m=8$ is similar.
(b) Binary maps $\to$ Distance maps $\to$ Direction maps.
The ground-truth binary maps are of category car, road and side-walk.
We first apply distance transform on each binary map to compute the
ground-truth distance maps.
Then we use Sobel filter on the distance maps to compute the 
ground-truth direction maps.
We choose different colors to represent different distance values 
or the direction values.
}
}
\label{fig:approach:angle2offset_gtdirection}
\vspace{-7mm}
\end{figure*}

We implement the direction branch as
$1\times 1~\operatorname{Conv} \rightarrow \operatorname{BN} \rightarrow \operatorname{ReLU}$ with $256$ output channels. 
We further apply a linear classifier ($1\times 1~\operatorname{Conv}$)
and up-sample the classifier prediction to generate the final direction map $\mathbf{D}$
of size $H\times W\times m$.
We mask the direction map $\mathbf{D}$ by multiplying
by the (binarized) boundary map $\mathbf{B}$ to ensure that we only apply direction 
loss on the pixels identified as boundary by the boundary branch.
We use the standard category-wise cross-entropy loss
to supervise the discrete directions in this branch.

% \vspace{.1cm}
\noindent\textbf{Offset branch.}
The offset branch is used to convert the predicted direction
map $\mathbf{D}$
to the offset map $\Delta{\mathbf{Q}}$ of size $H\times W\times 2$.
We illustrate the mapping mechanism in Figure~\ref{fig:approach:angle2offset_gtdirection} (a).
\textcolor{black}
{
For example, the ``upright'' direction category (corresponds to the
value within range $[0^{\circ}, 90^{\circ})$)
will be mapped to offset $(1, 1)$ when $m=4$.
}
Last, we generate the refined label map
through shifting the coarse label map 
with the grid-sample scheme~\cite{jaderberg2015spatial}.
The process is shown
in Figure~\ref{fig:approach:shift-improve}.

\vspace{-3mm}
\subsection{Ground-truth generation and analysis}
\label{sec:gtdir}

There may exist many different mechanisms to
generate ground-truth for the boundary maps
and the direction maps.
In this work, we mainly exploit the 
conventional distance transform~\cite{kimmel1996sub}
to generate ground-truth
for both semantic segmentation task and
the instance segmentation task.

% We mainly illustrate the details for semantic segmentation task.
We start from the ground-truth segmentation label
to generate the ground-truth distance map, followed by
boundary map and direction map.
Figure~\ref{fig:approach:angle2offset_gtdirection} (b) illustrates the overall procedure.

% \vspace{.1cm}
\noindent\textbf{Distance map.}
For each pixel,
our distance map records its 
minimum (Euclidean) distance to the pixels belonging to other object category.
We illustrate how to compute the distance map as below.

First, we decompose the ground-truth label into 
$K$ binary maps associated with different semantic categories,
e.g., car, road, sidewalk.
The $k^{th}$ binary map records the pixels belonging 
to the $k^{th}$ semantic category as $1$ and $0$ otherwise.
Second, we perform distance transform~\cite{kimmel1996sub}
\footnote{
We use \texttt{scipy.ndimage.morphology.distance$\_$transform$\_$edt}
in implementation.
} 
on each binary map independently to compute the distance map.
The element of $k^{th}$ distance map encodes the
distance from a pixel belonging to $k^{th}$ category
to the nearest pixel belonging to other categories.
Such distance can be treated as the distance to the object boundary. 
We compute a fused distance map through 
aggregating all the $K$ distance maps.

Note that the values in our distance map are (always positive)
different from the conventional signed distances that represent 
the interior/exterior pixels
with positive/negative distances separately.

% \vspace{.1cm}
\noindent\textbf{Boundary map.}
As the fused distance map represents the distances to the object boundary,
we can construct the ground-truth boundary map
through setting all the pixels with distance value smaller than a threshold
$\gamma$ as boundary\footnote{
We define the boundary pixels and interior pixels
based on their distance values.
}.
We empirically choose small $\gamma$ value, e.g., $\gamma=5$,
as we are mainly focused on the thin boundary refinement.

% \vspace{.1cm}
\noindent\textbf{Direction map.} 
We perform the Sobel filter (with kernel size $9\times 9$) 
on the $K$ distance maps independently to compute the 
corresponding $K$ direction maps respectively.
The Sobel filter based direction is in 
the range $[0^{\circ}, 360^{\circ})$,
and each direction points to the 
interior pixel (within the neighborhood) 
that is furthest away from the object boundary.
We divide the entire direction range to $m$ categories (or partitions) and 
then assign the direction of each pixel to the corresponding category.
We illustrate two kinds of partitions 
in Figure~\ref{fig:approach:angle2offset_gtdirection} (a)
and we choose the $m=8$ partition by default.
We apply the evenly divided direction map as our ground-truth for training.
Besides, we also visualize some examples of direction map in Figure~\ref{fig:approach:angle2offset_gtdirection} (b).

% \vspace{.1cm}
\noindent\textbf{Empirical Analysis.}
We apply the generated ground-truth
on the segmentation results of three state-of-the-art methods including DeepLabv3~\cite{chen2017rethinking}, HRNet~\cite{sun2019high} and Gated-SCNN~\cite{takikawa2019gated} 
to investigate the potential of our approach.
Specifically, 
we first project the ground-truth direction map to offset map
and then refine the segmentation results on Cityscapes \texttt{val} based on our generated ground-truth offset map.
Table~\ref{table:ideal_segfix} summarizes the related results. 
We can see that our approach significantly improves both the overall
mIoU and the boundary F-score.
% \footnote{The improvement ceiling of SegFix 
% depends on the performance of the segmentation model,
% i.e., the quality of the interior predictions.}.
For example, our approach ($m=8$) improves the mIoU of Gated-SCNN by $3.1\%$.
We may achieve higher performance through re-scaling 
the offsets for different pixels adaptively,
which is not the focus of this work.

% \vspace{.1cm}
\noindent\textbf{Discussion.}
The key condition for ensuring the effectiveness of our approach
is that \emph{segmentation predictions of the interior pixels are
more reliable empirically.}
Given accurate boundary maps and direction maps,
we could always improve the segmentation 
performance in expectation.
In other words, the segmentation performance ceiling 
of our approach is also determined by the interior pixels' prediction accuracy.

%##################################################################################################
\begin{table}[t]\centering
\footnotesize
\resizebox{0.75\linewidth}{!}
{
\tablestyle{5pt}{1.0}
\begin{tabular}{@{}l|c|c|cc@{}}
\shline
\multirow{2}{*}{metric} & \multirow{2}{*}{method} & \multirow{2}{*}{w/o SegFix} &\multicolumn{2}{c}{w/ SegFix} \\
% \cline{4-5}
& & & $m=4$ & $m=8$   \\
\shline
\multirow{3}{*}{mIoU} 
& DeepLabv3 (Our impl.)   & $79.5$ & $82.6$ (+$3.1$) & $82.4$ (+$2.9$) \\
& HRNet-W$48$ (Our impl.) & $81.1$ & $84.1$ (+$3.0$) & $84.1$ (+$3.0$)  \\
& Gated-SCNN  (Our impl.) & $81.0$ & $84.2$ (+$3.2$) & $84.1$ (+$3.1$)  \\
\hline
\multirow{3}{*}{F-score} 
& DeepLabv3 (Our impl.)   & $56.6$ & $68.6$ (+$12.0$) & $68.4$ (+$11.8$) \\
& HRNet-W$48$ (Our impl.) & $62.4$ & $73.8$ (+$11.4$) & $73.8$ (+$11.4$) \\
& Gated-SCNN (Our impl.)  & $61.4$ & $72.3$ (+$10.9$) & $72.3$ (+$10.9$) \\ 
\hline
\end{tabular}
}
\caption{
\small{
\textbf{
Improvements with ground-truth boundary offset on Cityscapes} \texttt{val}.
We report both the segmentation performance mIoU and the boundary performance F-score (1px width).
}
\label{table:ideal_segfix}}
\vspace{-9mm}
\end{table}

%%%%%%%%%%%%%%%%%%%%%%%%%%%%%%%%
\vspace{-2mm}
\section{Experiments: Semantic Segmentation}
\vspace{-2mm}
%%%%%%%%%%%%%%%%%%%%%%%%%%%%%%%%

\subsection{Datasets \& Implementation Details}

% \vspace{.1cm}
\noindent\textbf{Cityscapes}~\cite{cordts2016cityscapes}
is a real-world dataset that consists of $2,975$/$500$/$1,525$ images with resolution $2048\times 1024$ for training/validation/testing respectively. 
The dataset contains $19$/$8$ semantic categories for semantic/instance segmentation task.

% \vspace{.1cm}
\noindent\textbf{ADE20K}~\cite{zhou2017scene}
is a very challenging benchmark 
consisting of around $20,000$/$2,000$ images for training/validation respectively.
The dataset contains $150$ fine-grained semantic categories.

% \vspace{.1cm}
\noindent\textbf{GTA5}~\cite{richter2016playing}
is a synthetic 
dataset that consists of $12,402$/$6,347$/$6,155$ images with
resolution $1914\times 1052$ for training/validation/testing respectively.
The dataset contains $19$ semantic categories which are compatible with Cityscapes.

% \vspace{.1cm}
% \subsection{Experiment details}
\noindent\textbf{Implementation details}.
% \emph{training:}
We perform the same training and testing settings on Cityscapes and GTA5 
benchmarks as follow.
We set the initial learning rate as $0.04$, weight decay as $0.0005$, crop size as $512\times 512$ and batch size as $16$, and train for $80$K iterations.
For the ADE20K benchmark, we set the initial learning as $0.02$ and 
all the other settings are kept the same as on Cityscapes.
We use ``poly'' learning rate policy with power $=0.9$.
For data augmentation, we all apply random flipping horizontally, random cropping and random brightness jittering within the range of $[-10, 10]$.
Besides, we all apply syncBN~\cite{inplaceabn} across multiple GPUs to
stabilize the training.
We simply set the loss weight as $1.0$ for both the boundary loss and direction loss
without tuning.

Notably, our approach does not require extra training or fine-tuning any semantic segmentation models.
% \emph{
We only need to predict the boundary mask and the direction map 
for all the test images
\emph{in advance} and refine the segmentation results of any existing approaches accordingly.

\noindent\textbf{Evaluation metrics}.
We use two different metrics including:
\emph{mask F-score} and top-$1$ \emph{direction accuracy}
to evaluate the performance of our approach during the training stage.
Mask F-score is performed on 
the predicted binary boundary map and 
direction accuracy is performed on
the predicted direction map.
Especially, we only measure the direction accuracy within the regions identified as boundary by the boundary branch.

To verify the effectiveness of our approach for semantic segmentation,
we follow the recent Gated-SCNN~\cite{takikawa2019gated} and 
perform two quantitative measures including:
\emph{class-wise mIoU} to measure the overall segmentation performance on regions;
\emph{boundary F-score} to measure the boundary quality of predicted mask with a small slack in distance.
In our experiments,
we measure the boundary F-score using thresholds $0.0003$, $0.0006$ and $0.0009$ corresponding to $1$, $2$ and $3$ pixels respectively.
We mainly report the performance with threshold as $0.0003$ for most
of our ablation experiments.

% \vspace{-5mm}
\subsection{Ablation Experiments}
% \vspace{-2mm}
\label{sec:ablation}
We conduct a group of ablations to analyze the influence of various factors within SegFix.
We report the improvements over the segmentation baseline DeepLabv3 (mIoU/F-score
is $79.5\%$/$56.6\%$) if not specified.

\vspace{.1cm}
\noindent\textbf{Backbone.}
We study the performance of our SegFix based on
three different backbones with increasing complexities, i.e.,
HRNet-W$18$, HRNet-W$32$ and HRNet-${2\times}$.
We apply the same training/testing settings for all three backbones.
According to the comparisons in Table~\ref{table:segfix_backbone},
our SegFix consistently improves both the segmentation performance 
and the boundary quality with different backbone choices.
We choose HRNet-${2\times}$ in the following experiments 
if not specified as it performs best.
Besides, we also report their running time in Table~\ref{table:segfix_backbone}.

%##################################################################################################

\begin{table}[bpt]
\begin{minipage}[t]{1\linewidth}
\centering
\scriptsize
\resizebox{1\linewidth}{!}
{
\tablestyle{5pt}{1.0}
\begin{tabular}{@{}l|ccccccc@{}}
\shline
backbone         & $\#$param (M) &  runtime (ms) & {mask F-score} & {direction accuracy}  & mIoU$\triangle$ & F-score$\triangle$ \\\shline
HRNet-W18              & 9.6    & 16 & 71.44 & 64.44 & +0.8 & +3.7 \\
HRNet-W32              & 29.4   & 20 & 72.24 & 65.10 & +0.9 & +3.9 \\
HRNet-${2\times}$           & 47.3   & 69 & 73.67 & 66.87 & +1.0 & +4.4 \\
\hline
\end{tabular}
}
\caption{
\small{
\bd{Influence of backbones.}
The runtime is tested with an input image of resolution
$2048\times1024$ on a single V100 GPU (PyTorch1.4 + TensorRT).
SegFix reaches real-time speed with light-weight backbone, e.g., HRNet-W18 or HRNet-W32.
}
\label{table:segfix_backbone}}
\vspace{-2mm}
\end{minipage}
\begin{minipage}[t]{1\linewidth}
\centering
\scriptsize
\resizebox{0.8\linewidth}{!}
{
\tablestyle{5pt}{1.0}
\begin{tabular}{@{}l|cccc|ccc@{}}
\shline
& \multicolumn{4}{c|}{boundary width} & \multicolumn{3}{c}{\# directions} \\
& $\gamma=3$ & $\gamma=5$ & $\gamma=10$ & $\gamma=\infty$ & $m=4$ & $m=8$ & $m=16$\\\shline
mIoU$\triangle$           &  +0.94  & +0.96 & +0.95 & +0.84 &  +0.97  & +0.96 & +0.96  \\
F-score$\triangle$        &  +4.1  & +4.2  & +4.1 & +3.6 &  +4.1 &  +4.2 & +4.2  \\
\hline
\end{tabular}
}
\caption{
\small{
\bd{Influence of the boundary width and direction number.}
SegFix is robust to boundary width and direction number.
We choose $\gamma=5$ and $m=8$ according to their F-scores.
}
\label{table:segfix_boundary_width}}
\end{minipage}
\vspace{-5mm}
\end{table}

\begin{figure}
	\centering
	\begin{minipage}[c]{0.5\textwidth}
	\includegraphics[width=0.85\textwidth]{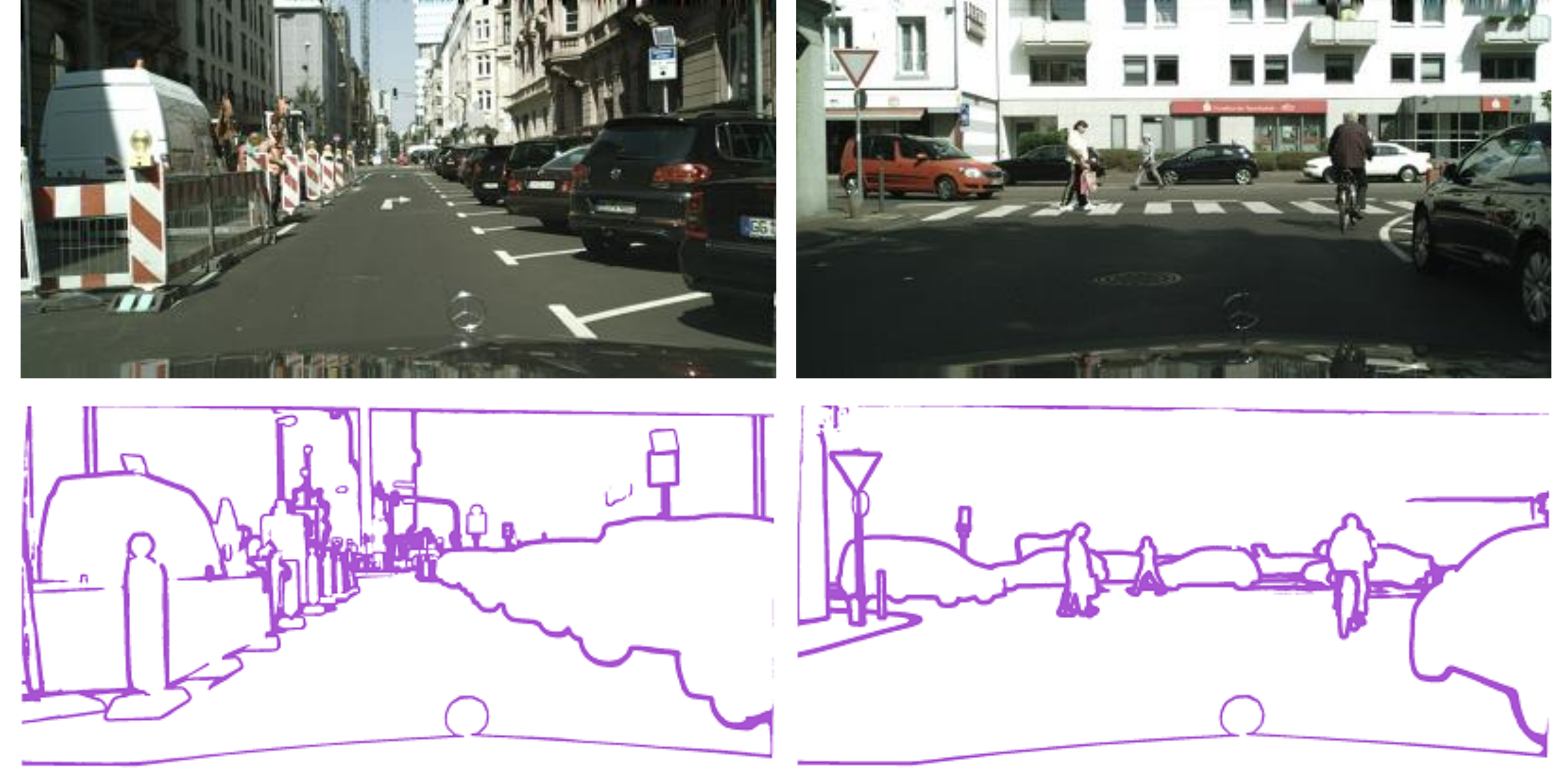}
	\end{minipage}
	\hspace{-7mm}
	\begin{minipage}[c]{0.4\textwidth}
	\caption{
	\small{
	\textbf{
	Qualitative results of our boundary branch prediction.
	}
	The $2$ example images are selected from Cityscapes \texttt{val}.
	We can see that their predicted boundaries are of high quality.
	}
	} \label{fig:boundary_map}
	\end{minipage}
	\vspace{-4mm}
\end{figure}

{
\vspace{-3mm}
\begin{figure*}
\centering
\resizebox{\linewidth}{!}
{
	\includegraphics[width=0.95\textwidth]{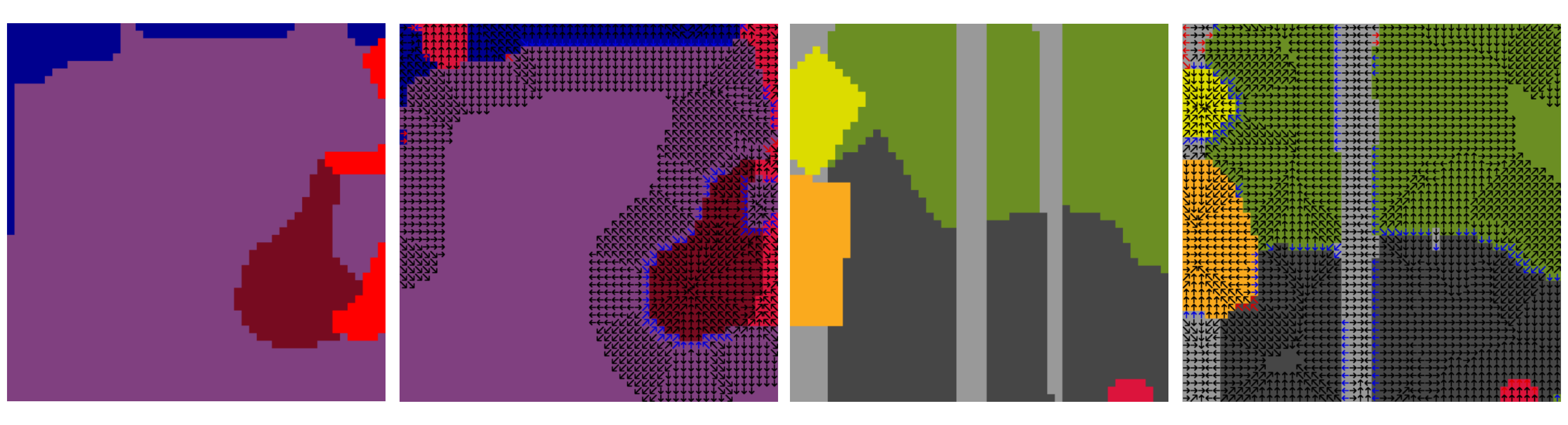}
}
\vspace{-6mm}
\caption{
\small{
\textbf{
Qualitative results of our direction branch predictions.
}
The \nth{1} and \nth{3} columns represent the ground-truth segmentation map.
The \nth{2} and \nth{4} columns illustrate the predicted directions 
with the segmentation map of HRNet as the background.
We mark the directions that fix errors with \textcolor{blue}{blue arrow} and directions that lead to extra errors with \textcolor{red}{red arrow}.
Our predicted directions addresses boundary errors for various object categories such as bicycle, traffic light and traffic sign.
(Better viewed zoom in)
}}
\label{fig:direction_example}
\vspace{-2mm}
\end{figure*}
}

\vspace{1mm}
\noindent\textbf{Boundary branch.}
We verify that SegFix is robust to
the choice of hyper-parameter $\gamma$ within the boundary branch 
and illustrate some qualitative results.

\noindent$\Box$ \textbf{boundary width}:
Table~\ref{table:segfix_boundary_width} shows the 
performance improvements based on boundary with different widths. 
We choose different $\gamma$ values to control the boundary width, 
where smaller $\gamma$ leads to thinner boundaries.
We also report the performance with $\gamma=\infty$,
which means all pixels is identified as boundary.
We find their improvements are close and we choose $\gamma=5$ by default.

\noindent$\Box$ \textbf{qualitative results}:
Figure~\ref{fig:boundary_map} shows 
the qualitative results
with our boundary branch.
We find that the predicted boundaries are
of high quality.
Besides, we also compute the F-scores between 
the boundary computed from the segmentation map of the
existing approaches, e.g., Gated-SCNN and HRNet, 
and the predicted boundary from 
our boundary branch.
The F-scores are
around $70\%$, which (in some degree) means
that their boundary maps are well aligned and 
ensures that more accurate direction
predictions bring larger performance gains.

\vspace{1mm}
\noindent\textbf{Direction branch.}
We analyze the influence of the direction number $m$ and 
then present some qualitative results of our predicted directions.

\vspace{1mm}
\noindent$\Box$ \textbf{direction number}:
We choose different direction numbers to perform different direction partitions
and control the generated offset maps that are used to refine the coarse label map.
We conduct the experiments with $m=4$, $m=8$ and $m=16$.
According to the reported results on the right $3$ columns in Table~\ref{table:segfix_boundary_width},
we find different direction numbers all lead to significant improvements
and we choose $m=8$ if not specified as our SegFix is 
less sensitive to the choice of $m$.

\vspace{1mm}
\noindent$\Box$ \textbf{qualitative results}:
In Figure~\ref{fig:direction_example},
we show some examples to illustrate that our predicted boundary directions
improve the errors.
Overall, the improved pixels (marked with \textcolor{blue}{blue arrow}) 
are mainly distributed along  the very thin boundary.

\vspace{1mm}
\noindent\textbf{Comparison with GUM.}
We compare SegFix with the previous model-dependent guided up-sampling mechanism~\cite{mazzini2018guided,Mazzini_2019_CVPR_Workshops} based on DeepLabv3 
as the baseline.
We report the related results in Table~\ref{table:compare_to_guide_upsample}.
It can be seen that our approach significantly outperforms GUM
measured by both mIoU and F-score.
We achieve higher performance through combining GUM with our approach,
which achieves $5.0\%$ improvements on F-score compared to the baseline.

\begin{table}[bpt]
\begin{minipage}[t]{1\linewidth}
\centering
\scriptsize
\resizebox{0.8\linewidth}{!}
{
\tablestyle{5pt}{1.0}
\begin{tabular}{@{}l|C{2cm}C{3cm}C{2cm}c@{}}
\shline
& baseline  &  GUM (Our impl.)  &  SegFix  &  GUM+SegFix   \\\shline
mIoU            & 79.5 & 79.8 (+0.3) & 80.5 (+1.0)  &  80.6 (+1.1) \\
F-score         & 56.6 & 57.7 (+1.1) & 60.9 (+4.3)  &  61.6 (+5.0)  \\
\hline
\end{tabular}
}
\caption{
\small{
\bd{Comparison with GUM}~\cite{mazzini2018guided}.
SegFix not only outperforms GUM but also is complementary with GUM.
}
\label{table:compare_to_guide_upsample}}
\end{minipage}
\begin{minipage}[t]{1\linewidth}
\centering
\scriptsize
\resizebox{0.8\linewidth}{!}
{
\tablestyle{5pt}{1.0}
\begin{tabular}{@{}l|C{2cm}C{2cm}C{2cm}c@{}}
\shline
& baseline & DenseCRF &  SegFix  &  DenseCRF+SegFix   \\\shline
mIoU            & 79.5 & 79.7 (+0.2) & 80.5 (+1.0) & 80.5 (+1.0) \\
F-score         & 56.6 & 60.9 (+4.3) & 61.0 (+4.4) & 64.1 (+7.5) \\
\hline
\end{tabular}
}
\caption{
\small{
\bd{Comparison with DenseCRF}~\cite{krahenbuhl2011efficient}.
SegFix achieves comparable F-score improvements and
much larger mIoU gains.
}}
\label{table:compare_to_crf}
\end{minipage}

\begin{minipage}[t]{0.46\linewidth}
	\centering
	\scriptsize
    \begin{tabular}{@{}l|cc@{}}
    \shline
     & Gated-SCNN &  Gated-SCNN+SegFix   \\\shline
    mIoU           & 81.0 & 81.5 (+0.5)  \\
    F-score        & 61.4 & 63.1 (+1.7)  \\
    \hline
    \end{tabular}
    \caption{
    \small{
    \bd{Comparison with Gated-SCNN}~\cite{takikawa2019gated}
    The result of Gated-SCNN is based on multi-scale testing.
    }
    \setlength{\tabcolsep}{5pt}
    }
    \label{table:compare_to_Gated-SCNN}
\end{minipage}
\hfill
\begin{minipage}[t]{0.5\linewidth}
	\centering
	\scriptsize
    \begin{tabular}{@{}l|cc|cc@{}}
    \shline
    & \multicolumn{2}{c|}{ADE20K} & \multicolumn{2}{c}{GTA5} \\
    & baseline & +SegFix & baseline & +SegFix \\
    \shline
    mIoU    & $44.8$ & $45.4 (+0.6)$ & $77.8$ & $80.6 (+2.8)$ \\
    F-score & $16.4$ & $19.3 (+2.9)$ & $50.2$ & $61.7 (+11.5)$  \\
    \hline
    \end{tabular}
    \caption{
    \small{
    \bd{DeepLabv3 with SegFix on ADE20K and GTA5.}
    We all choose DeepLabv3 as the baseline.
    }
    \label{table:gta5_synthia}
    }
\end{minipage}
\vspace{-5mm}
\end{table}

\vspace{1mm}
\noindent\textbf{Comparison with DenseCRF.}
We compare our approach with the conventional well-verified
DenseCRF~\cite{krahenbuhl2011efficient} based on 
the DeepLabv3 as our baseline.
We fine-tune the hyper-parameters of DenseCRF and set
them empirically following~\cite{chen2017deeplab}.
According to Table~\ref{table:compare_to_crf},
our approach not only outperforms DenseCRF
but also is complementary with DenseCRF.
The possible reasons for the limited mIoU improvements 
of DenseCRF might be that it brings more extra errors
on the interior pixels.

\vspace{1mm}
\noindent\textbf{Application to Gated-SCNN.}
Considering that Gated-SCNN~\cite{takikawa2019gated} 
introduced multiple components to 
improve the performance,
it is hard to compare our approach with Gated-SCNN fairly to a large extent.
To verify the effectiveness of our approach to some extent,
we first take the open-sourced Gated-SCNN (multi-scale testing) segmentation results
on Cityscapes validation set as the coarse segmentation maps,
then we apply the SegFix offset maps to refine the results.
We report the results in Table~\ref{table:compare_to_Gated-SCNN}
and SegFix improves the boundary F-score by $1.7\%$,
suggesting that SegFix is complementary with the strong
baseline that also focuses on improving the segmentation boundary quality.
Besides, we also report the detailed category-wise improvements measured by both mIoU and boundary F-score in Table~\ref{table:cityscapes_val}.

\vspace{1mm}
\noindent\textbf{Comparison with STEAL.}
Due to the training code of STEAL~\cite{acuna2019devil} is not open-sourced, 
we simply apply the released checkpoints\footnote{
\rm{STEAL:\space https://github.com/nv-tlabs/STEAL}
} to predict $K$ semantic boundary maps and convert them to binary boundary map. We empirically find that the boundary quality of our SegFix ($35.54\%$) is comparable with the carefully designed STEAL ($35.86\%$) measured by F-score along the ground-truth boundary with $1$-px width, suggesting that our method achieves nearly the state-of-the-art boundary detection performance.
To verify whether SegFix can benefit from the more accurate boundary maps predicted by STEAL,
we also train a SegFix model to only predict the direction map while using the (fixed) pre-computed boundary maps with STEAL.
We find the result becomes slightly worse ($80.5\%\to 80.32\%$) based on the coarse results with DeepLabv3.
% suggesting that the joint training of the boundary branch and direction branch might ensure better performance.

{
\begin{figure*}
\centering
\tiny{
\begin{tabularx}{\textwidth}{YYYYYY}
Image & Ground-Truth & DeepLabV3 & DeepLabV3+SegFix & HRNet & HRNet+SegFix \\
\end{tabularx}
}
\normalsize
{
\includegraphics[width=.158\textwidth]{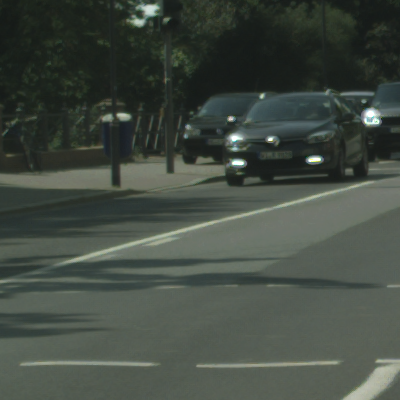}
\includegraphics[width=.158\textwidth]{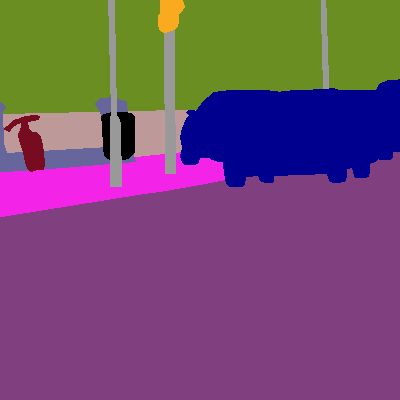}
\frame{\includegraphics[width=.158\textwidth]{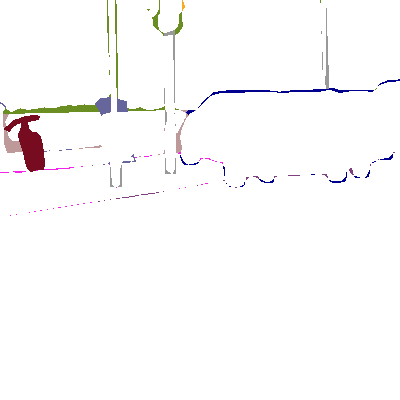}}
\frame{\includegraphics[width=.158\textwidth]{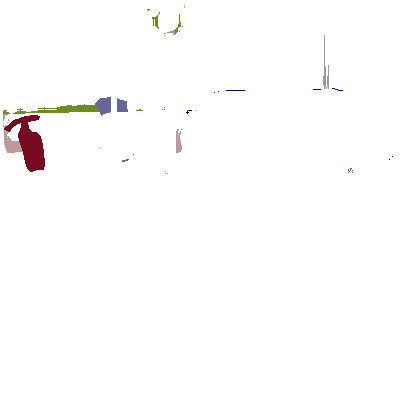}}
\frame{\includegraphics[width=.158\textwidth]{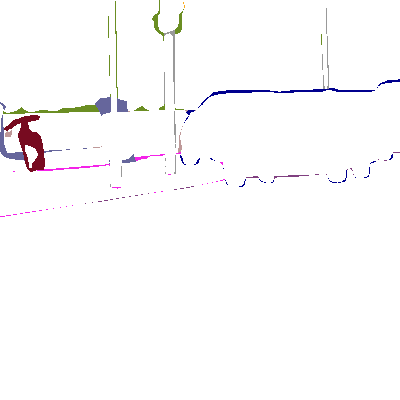}}
\frame{\includegraphics[width=.158\textwidth]{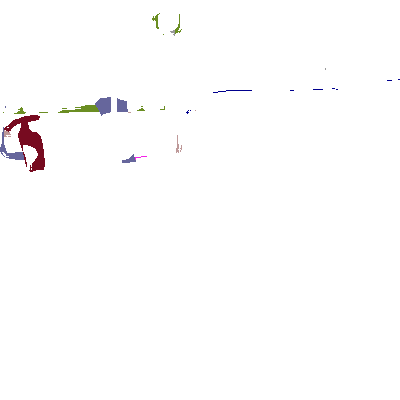}}
\\
\includegraphics[width=.158\textwidth]{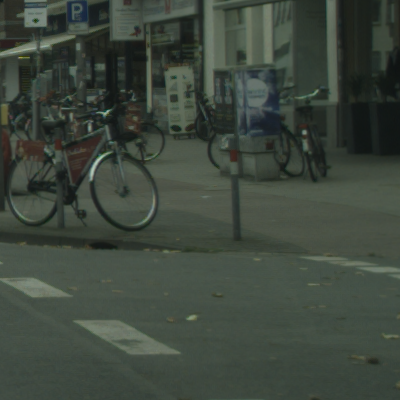}
\includegraphics[width=.158\textwidth]{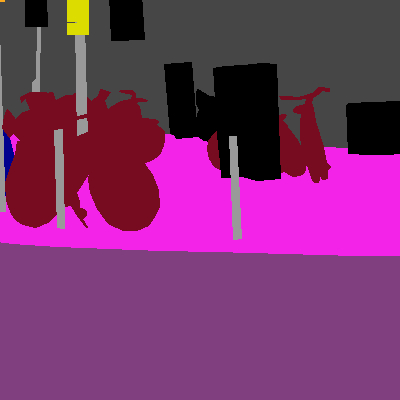}
\frame{\includegraphics[width=.158\textwidth]{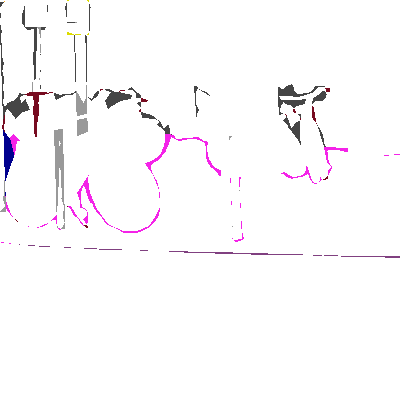}}
\frame{\includegraphics[width=.158\textwidth]{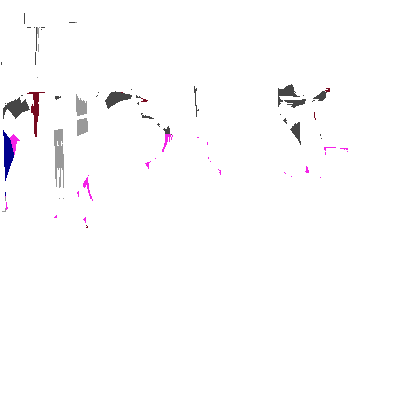}}
\frame{\includegraphics[width=.158\textwidth]{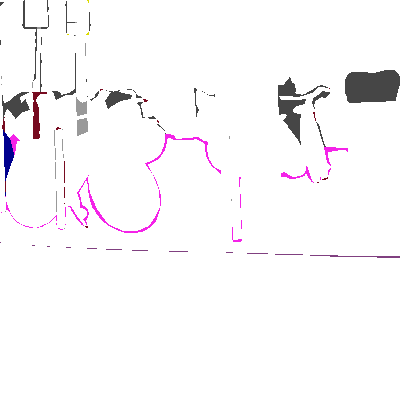}}
\frame{\includegraphics[width=.158\textwidth]{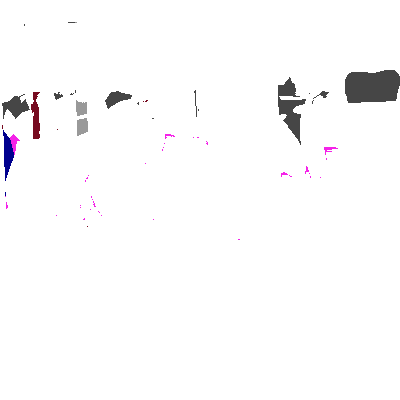}}
}
\vspace{-6mm}
\caption{\small{
\textbf{
Qualitative comparison in terms of errors on Cityscapes} \texttt{val}.
Our approach well addresses the existing boundary errors of various categories, e.g., car, bicycle and traffic sign, for both DeepLabv3 and HRNet.
}}
\label{fig:improve_hrnet}
\vspace{-3mm}
\end{figure*}
}

\vspace{1mm}
\subsection{Application to State-of-the-art}

We generate the boundary maps and the direction maps in advance and apply
them to the segmentation results of various state-of-the-art approaches 
without extra training or fine-tuning.

\vspace{1mm}
\noindent\textbf{Cityscapes val}:
We first apply our approach on various state-of-the-art approaches (on
Cityscapes \texttt{val}) including DeepLabv3, Gated-SCNN and HRNet.
We report the category-wise performance improvements in Table~\ref{table:cityscapes_val}.
It can be seen that our approach significantly improves the segmentation quality along the boundaries of all the evaluated approaches.
Figure~\ref{fig:improve_hrnet} provides some qualitative examples of
the improvements with our approach along the thin boundaries based on 
both DeepLabv3 and HRNet.

\vspace{1mm}
\noindent\textbf{Cityscapes test}:
We further apply our approach on several recent state-of-the-art methods on Cityscapes \texttt{test} including PSANet~\cite{zhao2018psanet}, DANet~\cite{fu2019dual}, BFP~\cite{ding2019boundary}, HRNet~\cite{sun2019high}, Gated-SCNN~\cite{takikawa2019gated}, VPLR~\cite{zhu2019improving}
and HRNet + OCR~\cite{yuan2019object}.
We directly apply the same model that are trained with only the $2,975$ training images without any other tricks, e.g., training with validation set or Mapillary Vistas~\cite{neuhold2017mapillary}, online hard example mining.

Notably, the state-of-the-art methods have applied various advanced techniques, e.g., multi-scale testing, multi-grid, performing boundary supervision or utilizing extra training data such as Mapillary Vistas or Cityscapes video, to improve their results.
In Table~\ref{table:segfix_sota},
our model-agnostic boundary refinement scheme consistently 
improves all the evaluated approaches. 
For example, with our SegFix, ''HRNet + OCR" achieves $84.5\%$ on Cityscapes \texttt{test}.
The improvements of our SegFix is in fact already significant
considering the baseline is already very strong and the performance gap between top ranking methods is just around $0.1\% \sim 0.3\%$.
We believe that lots of other advanced approaches might also benefit from our approach. 

% Notably, we only need to use SegFix to predict the offsets for once,
% and we apply the same offsets across different segmentation results.
% We also illustrate that we can train an unified SegFix model
% across different datasets in following Section~\ref{unify_segix}.

\begin{table}[bpt]
\begin{minipage}[t]{1\linewidth}
\centering
\scriptsize
\resizebox{\linewidth}{!}
{
\begin{tabular}{@{}l|l|ccccccccccccccccccc|c@{}} 
\shline
width & method & \rotatebox{90}{road} & \rotatebox{90}{sidewalk} & \rotatebox{90}{building} & \rotatebox{90}{wall} & \rotatebox{90}{fence} &\rotatebox{90}{pole} & \rotatebox{90}{traffic light}&\rotatebox{90}{traffic sign}&\rotatebox{90}{vegetation}&\rotatebox{90}{terrian}&\rotatebox{90}{sky}&\rotatebox{90}{person}&\rotatebox{90}{rider}&\rotatebox{90}{car}&\rotatebox{90}{truck}&\rotatebox{90}{bus} & \rotatebox{90}{train} & \rotatebox{90}{motorcycle} & \rotatebox{90}{bicycle} & mean\\\shline
\multirow{6}{*}{1px} & DeepLabV3 & 70.7 & 44.4 & 50.0 & 45.9 & 42.3 & 48.3 & 45.8 & 46.5 & 49.5 & 45.4 & 60.5 & 43.0 & 55.9 & 56.9 & 76.6 & 84.5 & 92.3 & 70.5 & 45.9 & 56.6 \\
& \textcolor{blue}{+ SegFix} & \textbf{73.9} & \textbf{49.1} & \textbf{55.5} & \textbf{47.8} & \textbf{43.7} & \textbf{57.6} & \textbf{52.7} & \textbf{58.3} & \textbf{54.7} & \textbf{47.4} & \textbf{64.7} & \textbf{50.2} & \textbf{59.7} & \textbf{64.6} & \textbf{77.4} & \textbf{86.0} & \textbf{92.6} & \textbf{72.0} & \textbf{51.5} & \textbf{61.0} \\
\cline{2-22}
& HRNet-W48 & 73.1 & 48.9 & 55.4 & 49.2 & 49.0 & 58.9 & 59.0 & 55.5 & 54.0 & 51.0 & 65.1 & 52.0 & 62.0 & 63.4 & 79.0 & 87.5 & 95.0 & 77.4 & 51.0 & 62.4 \\
& \textcolor{blue}{+ SegFix} & \textbf{74.8} & \textbf{51.9} & \textbf{58.2} & \textbf{50.9} & \textbf{49.7} & \textbf{63.6} & \textbf{64.0} & \textbf{61.6} & \textbf{57.1} & \textbf{52.5} & \textbf{66.8} & \textbf{56.8} & \textbf{64.4} & \textbf{67.5} & \textbf{79.7} & \textbf{88.7} & \textbf{95.2} & \textbf{77.7} & \textbf{55.0} & \textbf{65.1} \\
\cline{2-22}
& Gated-SCNN & 73.5 & 49.8 & 55.5 & 46.7 & 43.0 & 59.9 & 61.8 & 57.4 & 54.4 & 45.7 & 65.9 & 51.4 & 61.9 & 64.0 & 72.5 & 84.8 & 92.4 & 71.9 & 53.6 & 61.4 \\
& \textcolor{blue}{+ SegFix} & \textbf{74.2} & \textbf{51.3} & \textbf{57.7} & \textbf{47.2} & \textbf{45.3} & \textbf{64.0} & \textbf{63.8} & \textbf{61.2} & \textbf{56.7} & \textbf{46.9} & \textbf{66.6} & \textbf{55.6} & \textbf{64.0} & \textbf{66.9} & 72.0 & \textbf{85.0} & \textbf{92.6} & 71.8 & \textbf{55.9} & \textbf{63.1} \\
\hline
\multirow{6}{*}{2px} & DeepLabV3 & 79.1 & 57.5 & 62.2 & 49.3 & 45.5 & 64.1 & 54.5 & 61.3 & 62.6 & 49.8 & 72.2 & 54.8 & 62.4 & 71.6 & 78.0 & 86.5 & 92.7 & 72.3 & 54.7 & 64.8 \\
& \textcolor{blue}{+ SegFix} & \textbf{81.2} & \textbf{60.9} & \textbf{66.3} & \textbf{51.1} & \textbf{46.6} & \textbf{69.6} & \textbf{59.7} & \textbf{69.3} & \textbf{66.6} & \textbf{51.6} & \textbf{75.0} & \textbf{60.4} & \textbf{65.6} & \textbf{76.6} & \textbf{78.8} & \textbf{87.7} & \textbf{93.0} & \textbf{73.5} & \textbf{59.5} & \textbf{68.1} \\
\cline{2-22}
& HRNet-W48 & 81.1 & 61.7 & 67.4 & 52.5 & 52.5 & 73.2 & 67.7 & 69.4 & 66.9 & 55.4 & 76.3 & 63.7 & 68.2 & 77.3 & 80.4 & 89.6 & 95.5 & 79.1 & 60.3 & 70.4 \\
& \textcolor{blue}{+ SegFix} & \textbf{82.1} & \textbf{63.7} & \textbf{69.1} & \textbf{54.0} & \textbf{52.8} & \textbf{75.2} & \textbf{71.1} & \textbf{72.5} & \textbf{69.1} & \textbf{56.7} & \textbf{77.2} & \textbf{66.9} & \textbf{70.4} & \textbf{79.5} & \textbf{80.9} & \textbf{90.3} & \textbf{95.6} & 79.1 & \textbf{63.3} & \textbf{72.1} \\
\cline{2-22}
& Gated-SCNN & 80.9 & 61.9 & 67.1 & 50.0 & 46.4 & 73.9 & 70.3 & 70.1 & 67.1 & 50.0 & 76.7 & 62.8 & 68.5 & 77.3 & 74.0 & 86.8 & 92.9 & 73.8 & 62.5 & 69.1 \\
& \textcolor{blue}{+ SegFix} & \textbf{81.5} & \textbf{63.0} & \textbf{68.6} & \textbf{50.5} & \textbf{48.5} & \textbf{75.9} & \textbf{71.1} & \textbf{72.1} & \textbf{68.6} & \textbf{51.2} & \textbf{77.1} & \textbf{65.9} & \textbf{70.3} & \textbf{78.9} & 73.4 & 86.7 & \textbf{93.0} & 73.6 & \textbf{64.6} & \textbf{70.2} \\
\hline
\multirow{6}{*}{3px} & DeepLabV3 &84.1 & 65.8 & 70.7 & 52.0 & 47.9 & 72.5 & 60.8 & 70.2 & 72.2 & 53.2 & 79.9 & 62.9 & 67.3 & 79.8 & 79.0 & 87.8 & 93.0 & 73.7 & 61.6 & 70.2 \\
& \textcolor{blue}{+ SegFix} & \textbf{85.2} & \textbf{67.8} & \textbf{73.0} & \textbf{53.3} & \textbf{48.6} & \textbf{74.8} & \textbf{64.0} & \textbf{74.5} & \textbf{74.6} & \textbf{54.5} & \textbf{81.4} & \textbf{66.1} & \textbf{69.5} & \textbf{82.2} & \textbf{79.5} & \textbf{88.6} & \textbf{93.3} & \textbf{74.6} & \textbf{65.0} & \textbf{72.1} \\
\cline{2-22}
& HRNet-W48 & 85.5 & 69.1 & 74.7 & 54.9 & 54.9 & 79.0 & 72.9 & 75.6 & 75.5 & 58.6 & 83.0 & 70.4 & 72.6 & 84.3 & 81.3 & 90.8 & 95.7 & 80.3 & 66.9 & 75.1 \\
& \textcolor{blue}{+ SegFix} & \textbf{86.0} & \textbf{70.3} & \textbf{75.4} & \textbf{55.8} & 54.8 & \textbf{79.5} & \textbf{74.9} & \textbf{77.0} & \textbf{76.8} & \textbf{59.5} & \textbf{83.3} & \textbf{72.0} & \textbf{74.0} & \textbf{84.9} & \textbf{81.6} & \textbf{91.2} & \textbf{95.8} & 80.1 & \textbf{68.6} & \textbf{75.9} \\
\cline{2-22}
& Gated-SCNN & 85.0 & 68.8 & 74.2 & 52.2 & 48.7 & 79.7 & 75.0 & 75.9 & 75.4 & 53.0 & 83.1 & 69.3 & 73.1 & 83.6 & 74.9 & 87.8 & 93.2 & 75.2 & 68.8 & 73.5 \\
& \textcolor{blue}{+ SegFix} & \textbf{85.3} & \textbf{69.6} & \textbf{74.9} & \textbf{52.5} & \textbf{50.6} & \textbf{80.3} & 75.0 & \textbf{76.7} & \textbf{76.3} & \textbf{54.0} & \textbf{83.3} & \textbf{71.1} & \textbf{74.2} & \textbf{84.2} & 74.1 & 87.6 & 93.2 & 74.9 & \textbf{70.0} & \textbf{74.1} \\
\hline
\end{tabular}
}
\caption{
\small{
{
\textbf{Boundary F-score with SegFix.}
We illustrate the category-wise comparison with various baselines in terms of boundary F-score on Cityscapes \texttt{val}.
}
}
}
\label{table:cityscapes_val}
\end{minipage}
%%%%%%%%%%%%%%%%%%%%%%%%%%%%%%%%%%%%%%%%%%%%
\begin{minipage}[t]{1\linewidth}
\centering
\scriptsize
\resizebox{\linewidth}{!}
{
\tablestyle{5pt}{1.0}
\begin{tabular}{@{}l|ccccccccccccccccccc|c@{}} 
\shline
method & \rotatebox{90}{road} & \rotatebox{90}{sidewalk} & \rotatebox{90}{building} & \rotatebox{90}{wall} & \rotatebox{90}{fence} &\rotatebox{90}{pole} & \rotatebox{90}{traffic light}&\rotatebox{90}{traffic sign}&\rotatebox{90}{vegetation}&\rotatebox{90}{terrian}&\rotatebox{90}{sky}&\rotatebox{90}{person}&\rotatebox{90}{rider}&\rotatebox{90}{car}&\rotatebox{90}{truck}&\rotatebox{90}{bus} & \rotatebox{90}{train} & \rotatebox{90}{motorcycle} & \rotatebox{90}{bicycle} & mean\\ \shline
PSANet & 98.7 & 87.0 & 93.5 & 58.9 & 62.5 & 67.8 & 76.0 & 80.0 & 93.7 & 72.6 & 95.4 & 86.9 & 73.0 & 96.2 & 79.3 & 91.2 & 84.9 & 71.1 & 77.9 & 81.4  \\
\textcolor{blue}{+ SegFix} & 98.7 & \textbf{87.4} & \textbf{93.7} & \textbf{59.3} & \textbf{62.8} & \textbf{69.5} & \textbf{77.6} & \textbf{81.4} & \textbf{93.9} & \textbf{73.0} & \textbf{95.6} & \textbf{88.0} & \textbf{73.9} & \textbf{96.5} & \textbf{79.6} & \textbf{91.5} & \textbf{85.1} & \textbf{71.8} & \textbf{78.6} & \textbf{82.0} \\ 
\hline
DANet & 98.6 & 86.1 & 93.5 & 56.1 & 63.3 & 69.7 & 77.3 & 81.3 & 93.9 & 72.9 & 95.7 & 87.3 & 72.9 & 96.2 & 76.8 & 89.4 & 86.5 & 72.2 & 78.2 & 81.5 \\
\textcolor{blue}{+ SegFix} & \textbf{98.7} & \textbf{86.6} & \textbf{93.7} & \textbf{56.5} & \textbf{63.5} & \textbf{71.4} & \textbf{78.7} & \textbf{82.4} & \textbf{94.1} & \textbf{73.2} & \textbf{95.9} & \textbf{88.2} & \textbf{73.7} & \textbf{96.5} & \textbf{77.0} & \textbf{89.7} & \textbf{86.8} & \textbf{72.8} & \textbf{78.8} & \textbf{82.0} \\ 
\hline
BFP & 98.7 & 87.0 & 93.5 & 59.8 & 63.4 & 68.9 & 76.8 & 80.9 & 93.7 & 72.8 & 95.5 & 87.0 & 72.1 & 96.0 & 77.6 & 89.0 & 86.9 & 69.2 & 77.6 & 81.4 \\
\textcolor{blue}{+ SegFix} & 98.7 & \textbf{87.5} & \textbf{93.7} & \textbf{60.2} & \textbf{63.7} & \textbf{71.1} & \textbf{78.4} & \textbf{82.4} & \textbf{94.0} & \textbf{73.2} & \textbf{95.7} & \textbf{88.1} & \textbf{72.9} & \textbf{96.3} & \textbf{77.8} & \textbf{89.3} & \textbf{87.2} & \textbf{69.9} & \textbf{78.4} & \textbf{82.0} \\
\hline
HRNet & 98.8 & 87.5 & 93.7 & 55.6 & 62.3 & 71.8 & 79.3 & 81.8 & 94.0 & 73.1 & 95.8 & 88.5 & 76.1 & 96.5 & 72.2 & 86.5 & 84.7 & 73.8 & 79.4 & 81.8 \\
\textcolor{blue}{+ SegFix} & 98.8 & \textbf{87.9} & \textbf{93.9} & \textbf{56.0} & \textbf{62.5} & \textbf{73.6} & \textbf{80.7} & \textbf{83.2} & \textbf{94.1} & \textbf{73.4} & \textbf{95.9} & \textbf{89.3} & \textbf{76.7} & \textbf{96.6} & \textbf{72.4} & \textbf{86.7} & \textbf{85.0} & \textbf{74.3} & \textbf{80.2} & \textbf{82.2} \\ 
\hline
VPLR & 98.8 & 87.8 & 94.2 & 64.1 & 65.0 & 72.4 & 79.0 & 82.8 & 94.2 & 74.0 & 96.1 & 88.2 & 75.4 & 96.5 & 78.8 & 94.0 & 91.6 & 73.8 & 79.0 & 83.5 \\
\textcolor{blue}{+ SegFix} & 98.8 & \textbf{88.0} & \textbf{94.3} & \textbf{64.4} & \textbf{65.3} & \textbf{73.3} & \textbf{80.0} & \textbf{83.5} & \textbf{94.3} & \textbf{74.3} & \textbf{96.2} & \textbf{89.0} & \textbf{76.2} & \textbf{96.7} & \textbf{79.0} & \textbf{94.2} & \textbf{92.0} & \textbf{74.4} & \textbf{79.7} & \textbf{83.9} \\
\hline
HRNet + OCR & 98.9 & 88.3 & 94.3 & 66.8 & 66.6 & 73.6 & 80.3 & 83.7 & 94.3 & 74.4 & 96.0 & 88.7 & 75.4 & 96.6 & 82.5 & 94.0 & 90.8 & 73.8 & 79.7 & 84.2 \\
\textcolor{blue}{+ SegFix} & 98.9 & 88.3 & \textbf{94.4} & \textbf{68.0} & \textbf{67.8} & \textbf{73.6} & \textbf{80.6} & \textbf{83.9} & \textbf{94.4} & \textbf{74.5} & \textbf{96.1} & \textbf{89.2} & \textbf{75.9} & \textbf{96.8} & \textbf{83.6} & \textbf{94.2} & \textbf{91.3} & \textbf{74.0} & \textbf{80.1} & \textbf{84.5} \\
\hline
\end{tabular}
}
\caption{
\small{
\textbf{Segmentation mIoU with SegFix:}
{Category-wise improvements of SegFix based on various state-of-the-art methods on Cityscapes \texttt{test}.
Notably, ``HRNet + OCR + SegFix" ranks the first place on the Cityscapes
semantic segmentation leaderboard by the ECCV 2020 submission deadline.
% PSANet~\cite{zhao2018psanet}, DANet~\cite{fu2019dual}, BFP~\cite{ding2019boundary}
% HRNet~\cite{sun2019high}, VPLR~\cite{zhu2019improving} and HRNet + OCR~\cite{yuan2019object}.
}
}
\label{table:segfix_sota}}
\end{minipage}
%%%%%%%%%%%%%
\begin{minipage}[t]{1\linewidth}
\centering
\resizebox{0.7\linewidth}{!}
{
\tablestyle{5pt}{1.0}
\begin{tabular}{@{}l|cccccccc|l@{}} 
\shline
method & \rotatebox{90}{person}&\rotatebox{90}{rider}&\rotatebox{90}{car}&\rotatebox{90}{{truck}}&\rotatebox{90}{bus} & \rotatebox{90}{train} & \rotatebox{90}{motorcycle} & \rotatebox{90}{bicycle} & mean ($\%$) \\\shline
Mask-RCNN  & 36.0 & 28.8 & 51.6 & 30.0 & 38.7 & 27.3 & 23.9 & 19.4 & 32.0  \\
\textcolor{blue}{+ SegFix} & \bf{37.9} & \bf{30.3} & \bf{54.1} & \bf{31.0} & \bf{40.0} & \bf{27.9} & \bf{25.1} & \bf{20.5} & \bf{33.3} (+1.3)  \\
\hline
PointRend & 36.6 & 29.7 & 53.7 & 29.9 & 40.4 & 33.3 & 23.6 & 19.6 & 33.3 \\
\textcolor{blue}{+ SegFix} & \bf{38.7} & \bf{31.1} & \bf{56.2} & \bf{31.1} & \bf{41.6} & \bf{34.1} & \bf{24.6} & \bf{20.7} & \bf{34.8} (+1.5)\\ \hline
PANet  & 41.5 & 33.6 & 58.2 & 31.8 & 45.3 & 28.7 & 28.2 & 24.1 & 36.4 \\
\textcolor{blue}{+ SegFix} & \bf{43.3} & \bf{34.9} & \bf{60.4} & \bf{32.9} & \bf{47.0} & \bf{30.1} & \bf{29.1} & \bf{24.7} & \bf{37.8} (+1.4)\\ \hline
PolyTransform  & 42.4 & 34.8 & 58.5 & 39.8 & 50.0 & 41.3 & 30.9 & 23.4 & 40.1 \\
\textcolor{blue}{+ SegFix} & \bf{44.3} & \bf{35.9} & \bf{60.5} & \bf{40.5} & \bf{51.2} & \bf{41.6} & \bf{31.7} & \bf{24.1} & \bf{41.2} (+1.1)\\ 
\hline
\end{tabular}
}
\caption{
\small{
\bd{Results on Cityscapes Instance Segmentation task.}
Our SegFix significantly improves the mask AP
of Mask-RCNN~\cite{he2017mask}, PointRend~\cite{kirillov2019pointrend}, PANet~\cite{liu2018path}
and PolyTransform~\cite{liang2019polytransform} on Cityscapes \texttt{test} (w/ COCO pre-training).
Notably, ``PolyTransform + SegFix" ranks the second place on the Cityscapes
instance segmentation leaderboard by the ECCV 2020 submission deadline..
% The ${\bigstar}$ means the PointRend model is evaluated on Cityscapes \texttt{val}
% w/o COCO pre-training.
}
\label{table:segfix-instance-seg}}
\end{minipage}
\end{table}

%###############################################################################
\begin{table*}[htb]
\begin{minipage}[t]{0.9\linewidth}
\centering
\scriptsize
\resizebox{1\linewidth}{!}
{
\tablestyle{5pt}{1.0}
\begin{tabular}{@{}l|C{2cm}C{2cm}C{3cm}c@{}}
\shline
& DeepLabv3 & HRNet-W18 &  DeepLabv3+SegFix  &  DeepLabv3+HRNet-W18  \\\shline
mIoU            & 79.5 & 79.4  & 80.3 (+0.8) & 79.9 (+0.5) \\
F-score         & 56.6 & 57.0  & 60.3 (+3.7) & 58.2 (+1.6) \\
\hline
\end{tabular}
}
\caption{
\small{
\bd{Comparison with model ensemble.}
``DeepLabv3+HRNet-W18" reports the results based on model ensemble
and ``DeepLabv3+SegFix" reports the results based on our SegFix.
Our SegFix outperforms the model ensemble on both mIoU and F-score metrics.
We report the improvements compared to the performance with DeepLabv3.
}}
\label{table:compare_to_ensemble}
\end{minipage}
\vspace{-9mm}
\end{table*}

\vspace{-3mm}
\subsection{Experiments on ADE20K \& GTA5}
\vspace{-1.5mm}
We evaluate our SegFix scheme on two other challenging semantic segmentation 
benchmarks including ADE20K and GTA5.
We choose DeepLabv3
as our baseline on both datasets.
As illustrated in Table~\ref{table:gta5_synthia},
our approach also achieves significant performance
improvements along the boundary on both benchmarks,
e.g., the boundary F-score of DeepLabv3 gains $2.9\%$/$11.5\%$
on ADE20K \texttt{val}/GTA5 \texttt{test} separately.

%###############################################################################
\vspace{-3mm}
\subsection{Unified SegFix Model}
\vspace{-1.5mm}
\label{unify_segix}
% As boundary maps and direction maps are general across different datasets,
We propose to train a single unified SegFix model on Cityscapes and ADE20K,
and we  report the improvements over DeepLabv3 as below:
with a single unified SegFix model,
the performance gains are $0.9\%$/$3.8\%$ on Cityscapes
and $0.5\%$/$2.7\%$ on ADE20K measured by mIoU/F-score.
We can see these improvements are comparable with 
the SegFix trained on each dataset independently.
More experimental details are illustrated in the Appendix.

In general, we only need to train a single unified SegFix model
to improve the boundary quality of various segmentation 
models across different datasets,
thus SegFix is much more training friendly (and saves a lot of energy consumption) 
compared to the previous 
methods~\cite{bertasius2016semantic,takikawa2019gated,ding2019boundary,liu2018devil,liu2017learning,ke2018adaptive} that require re-training 
the existing segmentation models on each dataset independently.

\subsection{Comparison with Model Ensemble}
\label{sec:vs_ensemble}
To investigate whether our SegFix mainly benefits from model ensemble,
we conduct a group of experiments to compare our method with the standard model ensemble
(that ensembles two segmentation models with the same compacity)
under fair settings and report the results in Table~\ref{table:compare_to_ensemble}.
Specifically speaking, when processing a single image with resolution $1024\times2048$,
the overall computation cost of DeepLabv3+SegFix/DeepLabv3+HRNet-W18 is $2054$/$2060$ GFLOPs separately. 
We can see that SegFix outperforms the model ensemble,
e.g., DeepLabv3+SegFix gains $1.9\%$ (on F-score) over model ensemble method DeepLabv3+HRNet-W18,
suggesting that our SegFix is capable to fix that boundary errors
that the model ensemble fails to address.
Besides, another advantage of our method lies at that we can use a single unified SegFix model
across multiple datasets while the model ensemble requires training multiple different 
segmentation models on different datasets independently.
% Besides, we alo empirically verify that
% ``VPLR + SegFix" outperforms model ensemble
% ``VPLR + HRNet" by $0.3\%$ on Cityscapes \texttt{test},
% and this improvement is already significant considering the ``VPLR"~\cite{zhu2019improving}
% is already a very strong baseline.

% We illustrate the two main advantages of SegFix over model ensemble as following.
% First, we can train a unified SegFix model to improve the segmentation results
% across multiple datasets and our SegFix is agnostic of the previous
% segmentation models.
% However, we need to re-train multiple segmentation models
% for ensemble and the performance is also very sensitive to the ensemble weights.

% In our implementation, we use the same backbone HRNet-${2\times}$ for SegFix
% and we illustrate the training policy as below:
% we set the batch size as $16$ and construct each mini-batch by sampling $8$ images from Cityscapes
% and $8$ images from ADE20K.
% We choose the initial learning rate as $0.02$ and all the other training
% settings are kept the same.
% the same learning rate policy,
% the crop size as $512$ (for images from both datasets) and the same augmentation policy.

%%%%%%%%%%%%%%%%%%%%%%%%%%%%%%%%
\section{Experiments: Instance Segmentation}
%%%%%%%%%%%%%%%%%%%%%%%%%%%%%%%%

In Table~\ref{table:segfix-instance-seg},
we illustrate the results of SegFix on Cityscapes instance segmentation task.
We can find that the SegFix consistently improves the
mean AP scores over Mask-RCNN~\cite{he2017mask}, PANet~\cite{liu2018path},
PointRend~\cite{kirillov2019pointrend} and PolyTransform~\cite{liang2019polytransform}.
For example,
with SegFix scheme,
PANet gains $1.4\%$ points on the Cityscapes \texttt{test} set.
We also apply our SegFix on the very recent PointRend and PolyTransform.
Our SegFix consistently improves the performance of 
PointRend and PolyTransform 
by $1.5\%$ and $1.1\%$ separately,
which further verifies the effectiveness of our method.

We use the public available checkpoints from Dectectron2\footnote{
\rm{Detectron2: \space https://github.com/facebookresearch/detectron2}
}
and PANet\footnote{
\rm{PANet:\space https://github.com/ShuLiu1993/PANet}
} to generate the predictions of Mask-RCNN, PointRend and PANet.
Besides,
we use the segmentation results of PolyTransform directly. 
More training/testing details of SegFix on Cityscapes instance segmentation 
task are illustrated in the Appendix.
We believe that SegFix can be used to improve
various other state-of-the-art instance segmentation methods directly w/o any prior requirements.

Notably, the improvements on the instance segmentation tasks ($+1.1\%\sim1.5\%$) are more significant
than the improvements on semantic segmentation task ($+0.3\%\sim0.5\%$).
We guess the main reason is that the instance segmentation evaluation (on Cityscapes) only considers $8$ object categories without including the stuff categories. 
The performance of stuff categories is less sensitive to the boundary errors due to that their area is (typically) larger than the area of object categories. According to the category-wise results in Table 9, we can also find that the improvements on several object categories, e.g., person, rider, and truck, is more significant than the stuff categories, e.g., road, building.

%%%%%%%%%%%%%%%%%%%%%%%%%%%%%	
\vspace{-2mm}
\section{Conclusion}
\vspace{-2mm}
%%%%%%%%%%%%%%%%%%%%%%%%%%%%%
In this paper, 
we have proposed a novel model-agnostic approach
to refine the segmentation maps
predicted by an unknown segmentation model.
The insight is that 
the predictions of 
the interior pixels are more reliable.
We propose to replace the predictions of the boundary pixels
using the predictions of the corresponding interior pixels.
The correspondence is learnt only from the input image.
The main advantage of our method is that SegFix
generalizes well on various strong segmentation models.
Empirical results show that the effectiveness of our approach
for both semantic segmentation and instance segmentation tasks.
We hope our SegFix scheme can become a strong baseline
for more accurate segmentation results along the boundary.

\noindent \textbf{Acknowledgement:}
This work is partially supported by Natural Science Foundation of China under contract No. 61390511, and Frontier Science Key Research Project CAS No. QYZDJ-SSW-JSC009.

%%%%%%%%%%%%%%%%%%%%%%%%%%%%%
\section{Appendix}
%%%%%%%%%%%%%%%%%%%%%%%%%%%%%

First of all,
we need to clarify that all of our semantic segmentation ablation experiments
choose the DeepLabv3 as baseline if not specified.
% Second, we illustrate the effectiveness and failures of our approach
% with several more sections.
% In Section~\ref{sec:error_more_methods},
% we illustrate the statistics with more methods. 
In Section~\ref{sec:boundary_pixel},
we illustrate the statistics of the proportions of boundary pixels
over different categories as their scales vary so much.
In Section~\ref{sec:val_miou}, 
we report the category-wise mIoU improvements of our approach
on Cityscapes \texttt{val}.
In Section~\ref{sec:unified_segfix_model}, we provide
more details of our experiments with unified SegFix scheme.
In Section~\ref{sec:instance_details}, we present more details
of our experiments on Cityscapes instance segmentation task.
Last,
% in Section~\ref{sec:vs_ensemble}, we compare our SegFix to the model ensemble 
% scheme to verify the advantages of our method.
in Section~\ref{sec:vis_result}, we illustrate more qualitative 
results of our approach.

\begin{table*}[htb]
\begin{minipage}[t]{1\linewidth}
\centering
\scriptsize
\resizebox{\linewidth}{!}
{
\begin{tabular}{c|ccccccccccccccccccc|c} 
\toprule[1.2pt]
\pbox{10cm}{boundary width}
 & \rotatebox{90}{road} & \rotatebox{90}{{sidewalk}} & \rotatebox{90}{building} & \rotatebox{90}{{wall}} & \rotatebox{90}{{fence}} &\rotatebox{90}{pole} & \rotatebox{90}{traffic light}&\rotatebox{90}{traffic sign}&\rotatebox{90}{vegetation}&\rotatebox{90}{{terrain}}&\rotatebox{90}{sky}&\rotatebox{90}{person}&\rotatebox{90}{rider}&\rotatebox{90}{car}&\rotatebox{90}{{truck}}&\rotatebox{90}{bus} & \rotatebox{90}{train} & \rotatebox{90}{motorcycle} & \rotatebox{90}{bicycle} & mean\\\hline
1 & 1.0 & 4.2 & 2.7 & 4.4 & 4.3 & 20.1 & 12.2 & 9.0 & 2.8 & 5.2 & 3.5 & 8.1 & 10.9 & 2.5 & 2.2 & 2.4 & 3.2 & 8.4 & 8.3 & 2.6 \\
2 & 1.9 & 8.2 & 5.2 & 8.5 & 8.3 & 38.1 & 23.4 & 17.5 & 5.6 & 10.1 & 6.8 & 15.8 & 21.0 & 4.9 & 4.4 & 4.7 & 6.2 & 16.3 & 16.0 & 5.2 \\
3 & 2.7 & 11.4 & 7.2 & 12.0 & 11.7 & 51.6 & 32.3 & 24.2 & 7.8 & 14.0 & 9.5 & 21.6 & 28.2 & 6.8 & 6.2 & 6.7 & 8.6 & 21.9 & 21.4 & 7.2 \\
4 & 3.5 & 14.5 & 9.2 & 15.3 & 14.9 & 61.8 & 40.3 & 30.5 & 9.9 & 17.7 & 12.0 & 26.9 & 34.8 & 8.7 & 7.9 & 8.5 & 10.9 & 27.1 & 26.5 & 9.1 \\
5 & 4.4 & 18.0 & 11.4 & 18.8 & 18.2 & 69.8 & 48.5 & 37.1 & 12.4 & 21.7 & 15.0 & 33.4 & 42.6 & 11.0 & 9.8 & 10.7 & 13.6 & 33.2 & 32.4 & 11.2 \\
\bottomrule[0.8pt]
\end{tabular}
}
\caption{\small{The proportion of boundary pixels (with different widths) over different categories on Cityscapes \texttt{val} (\%).}}
\label{table:boundary_pixel_hist}
\end{minipage}
%%%%%%%%%%%%%%%%%
\begin{minipage}[t]{1\linewidth}
\centering
\scriptsize
\resizebox{\linewidth}{!}
{
\begin{tabular}{l|ccccccccccccccccccc|c} 
\toprule[1.2pt]
method & \rotatebox{90}{road} & \rotatebox{90}{sidewalk} & \rotatebox{90}{building} & \rotatebox{90}{wall} & \rotatebox{90}{fence} &\rotatebox{90}{pole} & \rotatebox{90}{traffic light}&\rotatebox{90}{traffic sign}&\rotatebox{90}{vegetation}&\rotatebox{90}{terrian}&\rotatebox{90}{sky}&\rotatebox{90}{person}&\rotatebox{90}{rider}&\rotatebox{90}{car}&\rotatebox{90}{truck}&\rotatebox{90}{bus} & \rotatebox{90}{train} & \rotatebox{90}{motorcycle} & \rotatebox{90}{bicycle} & mean\\
\toprule[1.2pt]
DeepLabV3 & 98.4 & 86.5 & 93.1 & 63.9 & 62.6 & 66.1 & 72.2 & 80.0 & 92.8 & 66.3 & 95.0 & 83.3 & 65.5 & 95.3 & 74.5 & 89.0 & 80.0 & 67.4 & 78.4 & 79.5 \\
\textcolor{blue}{+ SegFix} & \textbf{98.5} & \textbf{87.1} & \textbf{93.5} & \textbf{64.6} & \textbf{63.1} & \textbf{69.0} & \textbf{74.9} & \textbf{82.4} & \textbf{93.2} & \textbf{66.7} & \textbf{95.3} & \textbf{84.9} & \textbf{66.9} & \textbf{95.8} & \textbf{75.0} & \textbf{89.6} & \textbf{80.7} & \textbf{68.4} & \textbf{79.7} & \textbf{80.5} \\ 
\hline
HRNet-W48 & 98.5 & 87.0 & 93.5 & 58.5 & 64.7 & 71.4 & 75.6 & 82.8 & 93.2 & 64.8 & 95.3 & 84.7 & 66.9 & 95.8 & 82.9 & 91.5 & 82.9 & 69.8 & 80.1 & 81.1 \\
\textcolor{blue}{+ SegFix} & 98.5 & \textbf{87.4} & \textbf{93.7} & \textbf{59.0} & \textbf{65.1} & \textbf{72.5} & \textbf{77.0} & \textbf{84.0} & \textbf{93.4} & \textbf{65.1} & \textbf{95.4} & \textbf{85.7} & \textbf{67.7} & \textbf{96.1} & \textbf{83.1} & \textbf{91.9} & \textbf{83.4} & \textbf{70.8} & \textbf{81.0} & \textbf{81.6}\\ 
\hline
Gated-SCNN & 98.3 & 86.4 & 93.3 & 56.5 & 64.2 & 70.8 & 75.8 & 83.1 & 93.0 & 65.4 & 95.3 & 85.3 & 67.8 & 96.0 & 81.3 & 91.4 & 84.6 & 69.9 & 80.5 & 81.0 \\
\textcolor{blue}{+ SegFix} & \textbf{98.4} & \textbf{86.7} & \textbf{93.4} & \textbf{56.8} & \textbf{64.4} & \textbf{72.0} & \textbf{77.0} & \textbf{84.1} & \textbf{93.2} & \textbf{65.7} & \textbf{95.4} & \textbf{86.0} & \textbf{68.8} & \textbf{96.2} & \textbf{81.5} & \textbf{91.5} & \textbf{84.8} & \textbf{70.6} & \textbf{81.1} & \textbf{81.5} \\ 
\bottomrule[0.8pt]
\end{tabular}
}
\caption{\small{Category-wise mIoU improvements of SegFix based on various methods on Cityscapes \texttt{val}}.}
\label{table:segfix_val}
\end{minipage}
\vspace{-8mm}
\end{table*}

\subsection{Statistics of Boundary Pixels}
\label{sec:boundary_pixel}
We collect some statistics of the proportion of the boundary pixels
over different categories in Table~\ref{table:boundary_pixel_hist}.
We can find that the boundary pixels occupy large
proportions for three (small-scale) categories including
\emph{pole}, \emph{traffic light} and \emph{traffic sign}.
In fact, the performance improvements (measured by mIoU)
also mainly come from these three categories.
For example, in Table~\ref{table:segfix_val},
our SegFix improves the DeepLabv3's mIoUs of these three categories
by $3.1\%$, $2.7\%$ and $2.4\%$ separately.

\subsection{Category-wise mIoU Improvements}
\label{sec:val_miou}
We perform the SegFix on the 
Cityscapes \texttt{val} segmentation results
based on  DeepLabv3~\cite{chen2017deeplab}, 
Gated-SCNN~\cite{takikawa2019gated}
and HRNet~\cite{sun2019high}.
We report the category-wise mIoU improvements
in Table~\ref{table:segfix_val}
and we can see that our approach significantly
improves the performance on object categories including
\emph{pole}, \emph{traffic light} 
and \emph{traffic sign}.
The key reason might be that the objects 
belonging to
these categories tend to be of small scale,
which benefit more from the accurate boundary.

\subsection{Details of Unified SegFix Experiments}
\label{sec:unified_segfix_model}
In our implementation, we use the same backbone HRNet-${2\times}$ for SegFix
and we illustrate the training policy as below:
we set the batch size as $16$ and construct each mini-batch by sampling $8$ images from Cityscapes
and $8$ images from ADE20K.
We choose the initial learning rate as $0.02$ and all the other training
settings are kept the same.
the same learning rate policy,
the crop size as $512$ (for images from both datasets) and the same augmentation policy.
As illustrated in the paper, the performance of unified SegFix
is comparable with the performance of SegFix trained on each dataset separately.
In general, the proposed unified SegFix is a general scheme
that well addresses the boundary errors across multiple benchmarks.

\subsection{Details of Experiments on Instance Segmentation}
\label{sec:instance_details}

We generate the instance segmentation results of Mask-RCNN/PointRend
based on the open-sourced Detectron2~\cite{wu2019detectron2},
and we get the results of PANet~\cite{liu2018path} and PolyTransform~\cite{liang2019polytransform} from the authors directly
as our approach does not require training any segmentation models.

To predict suitable offset maps for instance segmentation,
we start from the instance masks and re-compute the ground-truth
distance maps, boundary maps and direction maps.
Specifically, for the instance pixels, we first estimate a distance map based on
each instance map
and then merge all the instance based distance maps as the final distance map.
We generate their direction maps and boundary maps following the same manner
as the manner for semantic segmentation.
% We also estimate the distance maps for the remained $11$ stuff categories.
% All the other training settings are kept the same,
We apply the predicted offset map on each predicted instance map
separately during the testing stage.
According to the experimental results on Cityscapes instance segmentation task,
we can see that SegFix consistently improves
the performance of various methods on Cityscapes \texttt{test}.
We also believe the recent state-of-the-art methods
might benefit from our SegFix.

\subsection{More Qualitative Results}
\label{sec:vis_result}
We illustrate more qualitative examples of the improvements (on semantic segmentation task)
with our approach in Figure~\ref{fig:improve_hrnet_supply}.
We can see that our approach well 
addresses the errors along thin boundary.
There still exist some errors located in the interior regions 
that our approach fail to address as we are mainly focused on 
the thin boundary refinement.

{
\begin{figure*}
\centering
\tiny{
\begin{tabularx}{\textwidth}{YYYYYY}
  {Image} & {Ground-Truth} & {DeepLabV3} & {DeepLabV3+SegFix} & {HRNet} & {HRNet+SegFix}
\end{tabularx}
}
\normalsize
{
	\includegraphics[width=.158\textwidth]{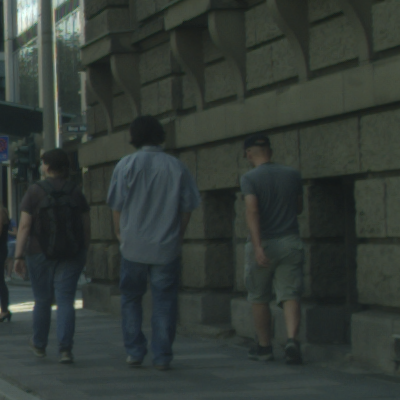}
	\includegraphics[width=.158\textwidth]{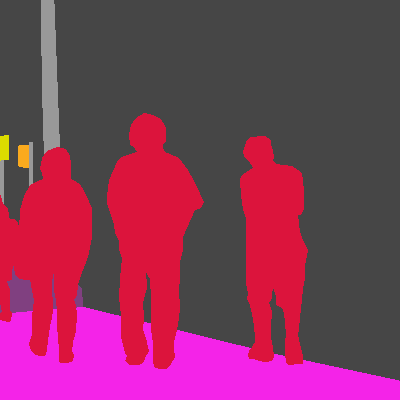}
	\frame{\includegraphics[width=.158\textwidth]{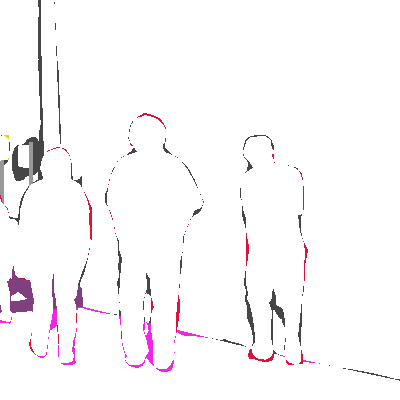}}
	\frame{\includegraphics[width=.158\textwidth]{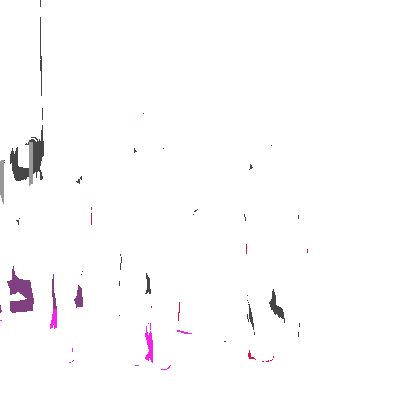}}
	\frame{\includegraphics[width=.158\textwidth]{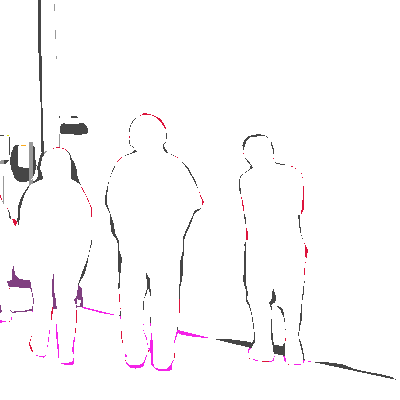}}
	\frame{\includegraphics[width=.158\textwidth]{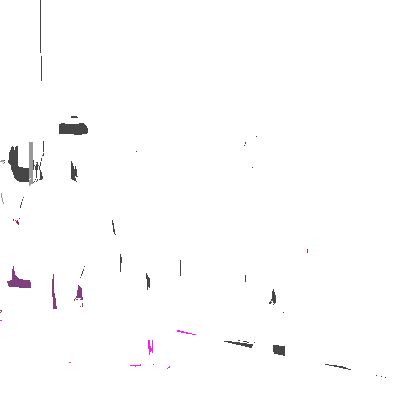}}
	\\
	\vspace{.05cm}
	\includegraphics[width=.158\textwidth]{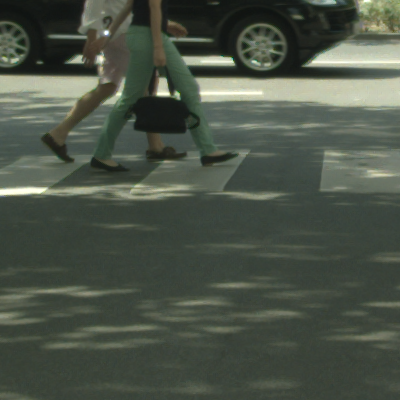}
	\includegraphics[width=.158\textwidth]{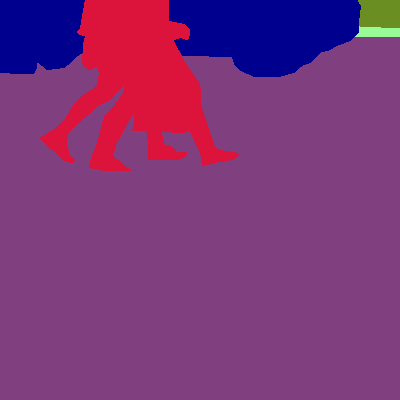}
	\frame{\includegraphics[width=.158\textwidth]{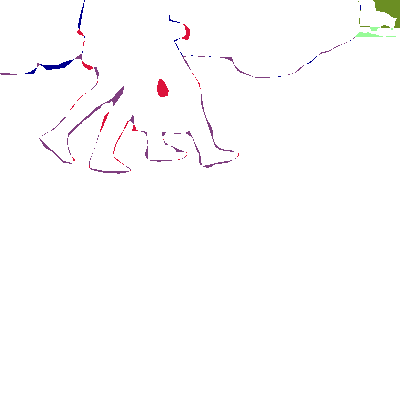}}
	\frame{\includegraphics[width=.158\textwidth]{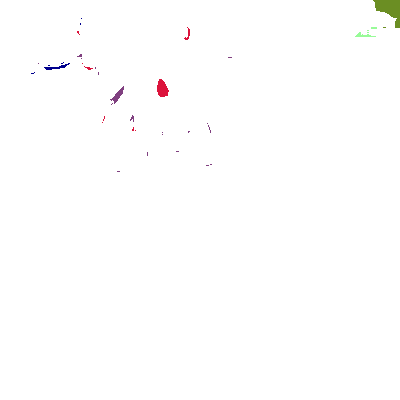}}
	\frame{\includegraphics[width=.158\textwidth]{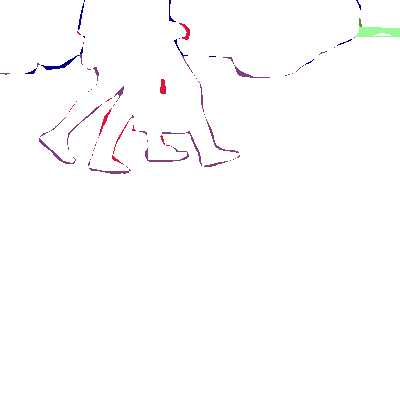}}
	\frame{\includegraphics[width=.158\textwidth]{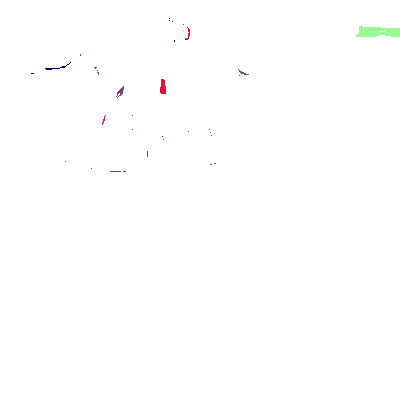}}
	\\
	\vspace{.05cm}
	\includegraphics[width=.158\textwidth]{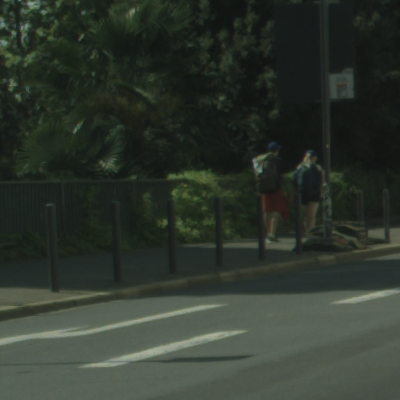}
	\includegraphics[width=.158\textwidth]{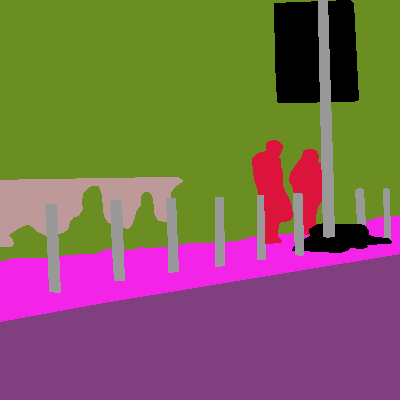}
	\frame{\includegraphics[width=.158\textwidth]{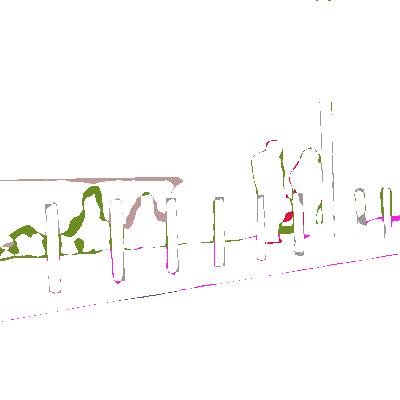}}
	\frame{\includegraphics[width=.158\textwidth]{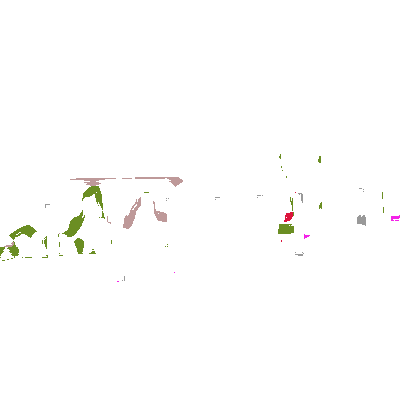}}
	\frame{\includegraphics[width=.158\textwidth]{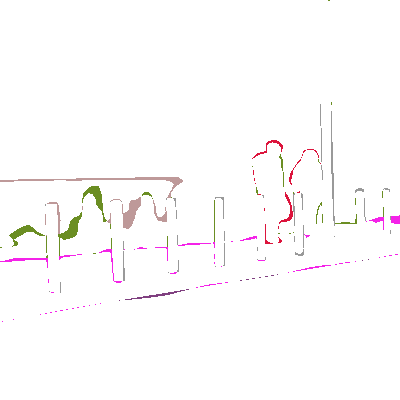}}
	\frame{\includegraphics[width=.158\textwidth]{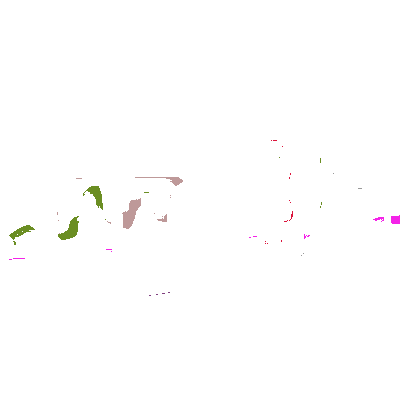}}
	\\
	\vspace{.05cm}
	\includegraphics[width=.158\textwidth]{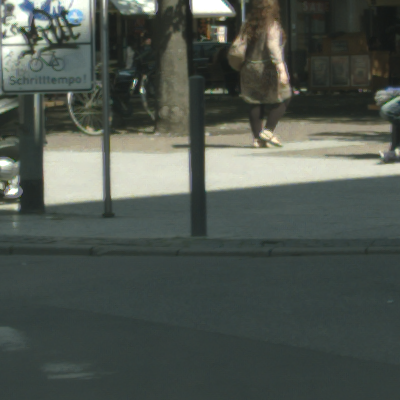}
	\includegraphics[width=.158\textwidth]{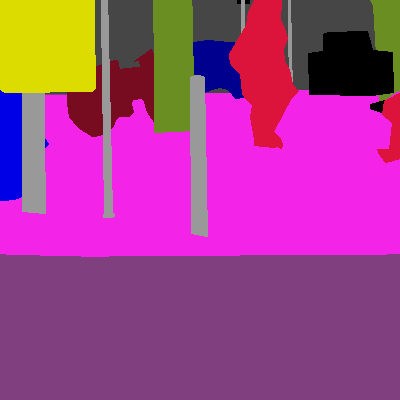}
	\frame{\includegraphics[width=.158\textwidth]{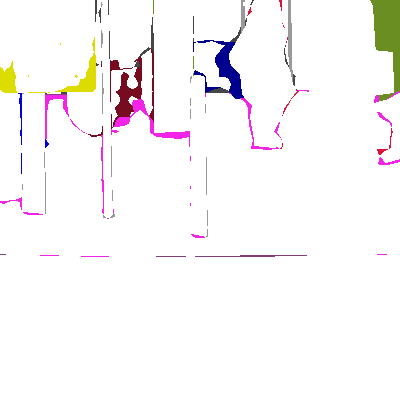}}
	\frame{\includegraphics[width=.158\textwidth]{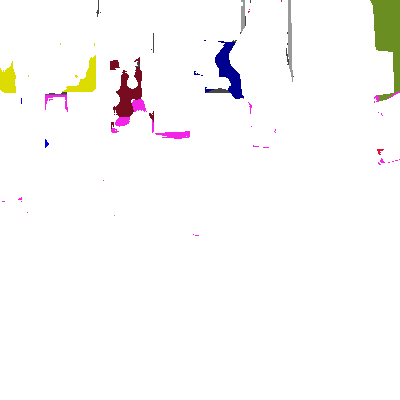}}
	\frame{\includegraphics[width=.158\textwidth]{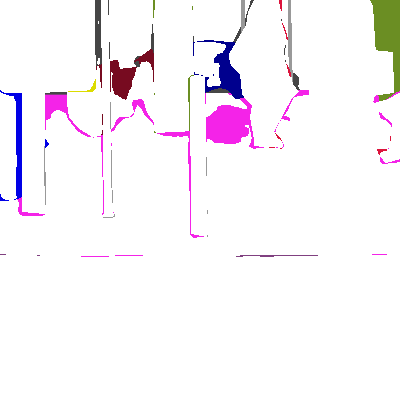}}
	\frame{\includegraphics[width=.158\textwidth]{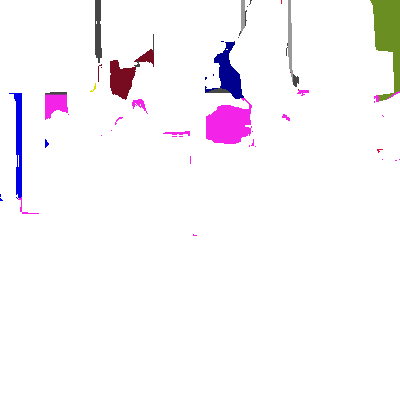}}
	\\
	\vspace{.05cm}
	\includegraphics[width=.158\textwidth]{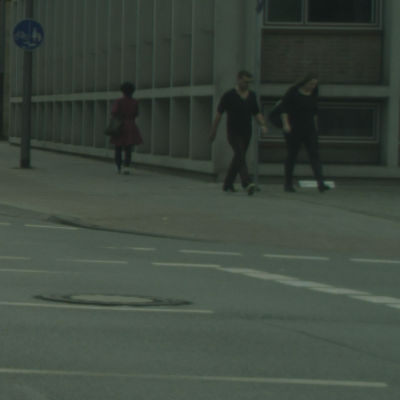}
	\includegraphics[width=.158\textwidth]{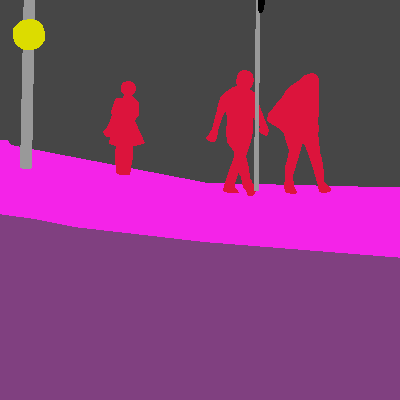}
	\frame{\includegraphics[width=.158\textwidth]{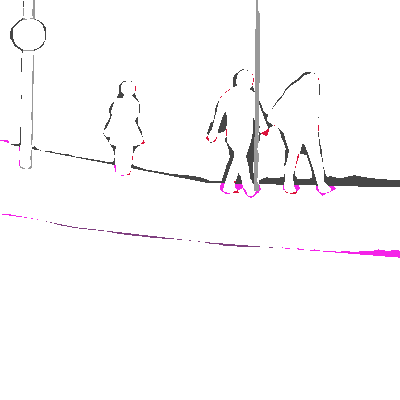}}
	\frame{\includegraphics[width=.158\textwidth]{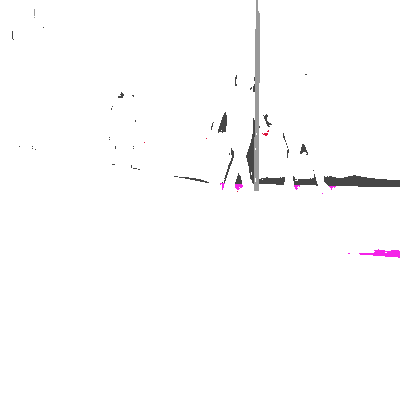}}
	\frame{\includegraphics[width=.158\textwidth]{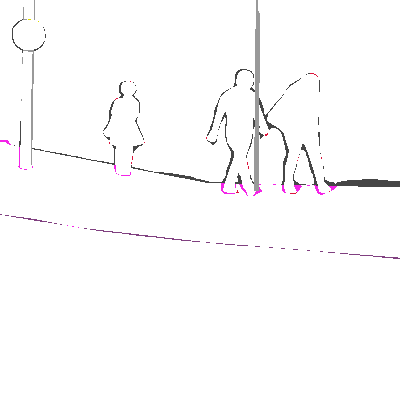}}
	\frame{\includegraphics[width=.160\textwidth]{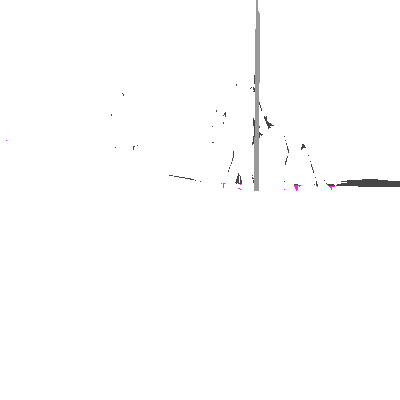}}
}
\vspace{-0.5cm}
\caption{\small{
Qualitative comparison in terms of errors on Cityscapes \texttt{val}.
Our approach well addresses the existing boundary errors of various categories,
e.g., person, pole and traffic sign, for both DeepLabv3 and HRNet.
}}
\label{fig:improve_hrnet_supply}
\end{figure*}
}

\bibliographystyle{splncs04}\small
\bibliography{segfix}
\end{document}